\documentclass{article} 
\usepackage{iclr2027_conference,times}

\usepackage[utf8]{inputenc}
\usepackage[T1]{fontenc}
\usepackage{microtype}

\usepackage{amsmath, amssymb, amscd, amsfonts, mathtools}
\usepackage{dsfont}
\usepackage{bbm}
\usepackage{bm}
\usepackage{physics}
\usepackage{cancel}
\usepackage{tensor}
\usepackage{latexsym}
\usepackage{nicefrac}
\usepackage{bigints}

\usepackage{booktabs}
\usepackage{multirow}
\usepackage{makecell}
\usepackage{floatrow}
\usepackage{caption}
\usepackage[caption=false]{subfig}
\usepackage{float}
\usepackage{placeins}
\usepackage{rotating}
\newlength{\tabw}
\usepackage{graphicx}
\usepackage{epsfig}
\usepackage{epstopdf}
\usepackage{xcolor}
\usepackage{color}

\usepackage{tikz}
\usetikzlibrary{matrix}
\usetikzlibrary{decorations.markings, calc, shapes, decorations.pathmorphing, patterns, decorations.pathreplacing, arrows.meta}
\usetikzlibrary{positioning, fit, backgrounds}

\usepackage{tcolorbox}

\usepackage[toc,page]{appendix}

\usepackage{url}
\usepackage{hyperref}
\hypersetup{
 colorlinks=true,
 citecolor=blue,
 linkcolor=blue,
 urlcolor=blue
}

\DeclareTextFontCommand{\ba}{\bfseries\sffamily}

\def\vecr{\mathbf{r}}
\def\vecR{\mathbf{R}}

\title{Hessian Matching for Machine-Learned Coarse-Grained Molecular Dynamics}

\author{
\makebox[\textwidth]{\hfill \textbf{Sanya Murdeshwar}$^{1}$ \hspace{0.3cm} \textbf{Sanjit Shashi}$^{1,2}$ \hfill} \\[0.3em]
\makebox[\textwidth]{\textbf{Kevin Bachelor}$^{1}$ \hspace{0.3cm} \textbf{William Noid}$^{3}$ \hspace{0.3cm} \textbf{Razvan Marinescu}$^{1,2}$} \\[0.3em]
\small \mbox{$^{1}$University of California, Santa Cruz} \quad
 \mbox{$^{2}$GiwoTech Inc.} \quad
 \mbox{$^{3}$Pennsylvania State University} \\[0.3em]
\small \texttt{\{smurdesh, sashashi, kwbachelor, ramarine\}@ucsc.edu},\hspace{0.3cm} \small \texttt{wgn1@psu.edu}
}

\iclrfinalcopy 
\begin{document}

\maketitle
\lhead{Preprint}   

\begin{abstract}
Coarse-grained (CG) molecular dynamics enables simulations of atomic systems such as biomolecules at timescales inaccessible to all-atom (AA) methods, but existing CG neural potentials trained via force matching capture only the gradient of the free-energy surface, leaving its curvature unsupervised. We introduce a framework that augments force matching with stochastic Hessian-vector product (HVP) matching, instilling second-order curvature information into CG potentials without constructing the full Hessian. We derive a decomposition of the target CG Hessian into a model-independent projected AA Hessian, precomputed once before training, and a model-dependent covariance correction computed online at negligible cost. We then construct an unbiased stochastic estimator of the Hessian-matching objective by using random probe vectors. We evaluate our method by training on a benchmark set of nine fast-folding proteins and gauging agreement of the learned free-energy landscapes with reference data along the directions of slowest collective motion using time-lagged independent component analysis. HVP matching improves slow-mode accuracy over force matching for all but one of our benchmark proteins. It also sharpens local structure, reducing bond-length error on eight of nine proteins, and in several cases by an order of magnitude. Our results demonstrate that higher-order physical supervision is a practical path to more accurate CG potentials for biomolecular simulation.
\end{abstract}

\section{Introduction}

Molecular dynamics (MD) simulations are a primary tool for studying the dynamical and functional properties of biomolecules. However, the timescales accessible to all-atom (AA) MD remain fundamentally mismatched with the biological processes that researchers aim to study. Certain meaningful processes, such as protein folding, can take milliseconds to seconds, which remains costly or out of reach for AA MD even on specialized hardware~\citep{Shaw2009,Shaw2021}. This gap motivates coarse-grained (CG) modeling~\citep{Noid2013,Kmiecik2016}. CG models represent groups of atoms as single interaction sites (beads), reducing the number of degrees of freedom and enabling simulations of longer timescales. The central challenge is learning a CG potential whose free-energy surface reproduces that of the underlying AA system, including the locations of metastable states, the heights of barriers between them, and the local curvature that governs fluctuations within each basin.

CG models have traditionally been parameterized top-down to reproduce experimental observables, as in the Martini model~\citep{Marrink2007}. Bottom-up approaches instead derive CG interactions from AA simulations, either from structural distributions~\citep{Lyubartsev1995,Tschop1998,Akkermans2001}, as in iterative Boltzmann inversion~\citep{MullerPlathe2002}, or from forces, as in force matching~\citep{Ercolessi1994,Izvekov2005,Noid2008}, which fits the CG potential to the mean force of the AA reference. Force matching has become the dominant approach for machine-learned CG potentials such as CGnet~\citep{Wang2019} and CGSchNet~\citep{Husic2020}, which can produce stable, reasonably accurate simulations of fast-folding proteins.

Despite this progress, a key limitation of force matching remains. It only supervises the gradient of the free-energy surface, not its curvature. With finite training data, two potentials can agree on the forces at every sampled configuration while differing in how those forces vary around them, and force matching cannot distinguish between the two. Force matching also has known difficulty recovering global properties of the free-energy surface, such as the relative populations of metastable basins~\citep{Thaler2022, Khler2023, Noid2024}, with the FM validation loss failing to reflect global free-energy surface quality~\citep{Thaler2022}. These limitations can also manifest in out-of-distribution regimes, where models trained on one set of proteins extrapolate poorly to unseen sequences, producing unrealistically low energies in configurations not sampled during training~\citep{Majewski2023}. We examine this setting separately in Appendix~\ref{app:generalization}. These issues fundamentally limit the data efficiency and stability of CG neural potentials.

A natural approach to supplying the missing curvature information is to match the Hessian of the free-energy surface alongside its gradient (forces). The Hessian encodes higher-order information about local curvature, which governs vibrational frequencies and basin shapes. However, directly incorporating Hessian supervision into training CG neural potentials imposes major computational challenges with regards to scaling. For a CG system with $d = 3N$ degrees of freedom, the full Hessian has $d^2$ entries and requires $\mathcal{O}(d)$ force evaluations to construct column-by-column, in contrast to the gradient, which only has $d$ entries and requires $\mathcal{O}(1)$ force evaluations. For small systems, this is feasible but slow. For larger systems, $d$ scales into the thousands, and explicit construction becomes intractable both in memory and compute.

\subsection{Related work}
\label{sec:related}

Several recent works have incorporated second-order curvature information into atomistic machine-learned interatomic potentials (MLIPs), showing improvements in extrapolation, data efficiency, and transition state optimization \citep{Fang2024BeyondNumerical,Yuan2024AnalyticalHessian,Rodriguez2025HessianData}. The most closely related proposal to ours in this vein is Projected Hessian Learning (PHL) \citep{Rodriguez2026PHL}, which replaces explicit Hessian construction with stochastic Hessian-vector product probes. PHL produces an unbiased trace-based loss that recovers most of the benefits of full Hessian supervision at under 8\% of the cost. Hessian Interatomic Potentials (HIP) \citep{HIP2025} takes a different route, directly predicting Hessians with an equivariant readout rather than via autograd.

However, all of these methods apply to atomistic MLIPs trained on exact quantum-mechanical Hessians of very small molecules. CG systems can represent much larger molecules, making stochastic HVP probes essential rather than optional. Furthermore, the CG Hessian decomposes into a projected AA Hessian plus a covariance correction arising from integrated-out atomic degrees of freedom, and the latter has no analogue in the atomistic regime. We also note that one of the goals in CG modeling is generalization to unseen protein conformations and sequences, and no existing Hessian-informed methods have been evaluated to that end.

A parallel line of work pursues transferable CG protein potentials via diverse multi-protein training. \citet{Majewski2023} trained CG potentials on twelve proteins jointly, and \citet{Charron2025Navigating} developed a transferable CGSchNet-based force field demonstrating extrapolative MD on new sequences. Both rely solely on force matching, achieving transferability through data scale rather than richer physical supervision.

The multiscale coarse-graining formalism~\citep{Noid2008} established the theoretical basis for force matching with general (linear and nonlinear) CG maps. The associated mean-force relations can be derived via the vectorial Blue Moon ensemble~\citep{Ciccotti2005} and were reformulated as a geometric projection problem by \citet{Kalligiannaki2015}, who also showed that force matching and relative-entropy minimization are asymptotically equivalent for general CG maps. Our work extends this framework from first to second derivatives of the free energy.

Our work is also distinct from optimizer-side curvature methods such as L-BFGS, K-FAC, and Hessian-free optimization~\citep{Martens2010, MartensGrosse2015, Pearlmutter1994}, which approximate the Hessian of the training objective in parameter space rather than in configuration space. 

\newpage
\paragraph{Our contribution.}
We introduce a framework for training CG neural potentials with stochastic Hessian matching via Hessian-vector products (HVPs). We demonstrate that the CG Hessian identity decomposes cleanly into two terms with fundamentally different computational properties: a projected AA Hessian term that is model-independent and can be precomputed once before training, and a covariance correction term that is model-dependent but built from quantities already available during the forward pass. We match the action of this Hessian, a $d \times d$ matrix, on $K$ random probe vectors, avoiding constructing the entire Hessian explicitly, all while retaining unbiased estimation of the full Hessian-matching objective. The resulting loss contributes $\mathcal{O}(Kd)$ additional work per frame for $K$ probes, which is linear in system size, and requires no architectural changes to the underlying model.

Our experimental results on fast-folding benchmark proteins show that supplementing force matching with HVP matching provides improved recovery of the slow collective modes (the time-lagged independent components capturing folding/unfolding transitions) on the majority of benchmark proteins. We also find improved alignment with several aspects of ground-truth molecular structure, particularly bond lengths, bond angles, and radii of gyration, for most proteins. Ultimately, these findings suggest that instilling higher-order physical information into CG neural potentials is a practical, scalable path to more accurate biomolecular simulation.

\section{Deriving the coarse-grained Hessian identity}
\label{sec:derivation}
 
We derive the symbolic expression for the CG Hessian of the free-energy surface, which serves as the training objective for our model. For proteins, the standard coarse-graining mechanism places one bead at the C$_\alpha$ atom of each amino acid residue. Note that because each bead is selected from the AA atoms, the mapping is linear, simplifying our derivation significantly. The general nonlinear expression is found in Appendix \ref{app:genBlueMoon}.
 
We make use of the Blue Moon ensemble~\citep{Carter1989,Kidder2021} to compute thermodynamic quantities associated with constrained molecular systems. Suppose that the AA system contains $n$ atoms and that the CG system consists of $N < n$ beads, both in three dimensions. The partition function is
\begin{equation}
Z(\vecR) = \int d\vecr\,\delta(\boldsymbol{\Xi}_r \vecr - \vecR) e^{-\beta \mathcal{H}(\vecr)},\ \ \ \ \beta \equiv \frac{1}{k_{\text{B}}T}.
\end{equation}
Here $\vecr \in \mathbb{R}^{3n}$ is an element of the full configuration space, $\boldsymbol{\Xi}_r$ is the $3N \times 3n$ coarse-graining matrix mapping AA positions to CG positions, and $\vecR \in \mathbb{R}^{3N}$ is an element of the CG configuration space. The Hamiltonian $\mathcal{H}$ is a scalar function. $\beta$ is the thermodynamic parameter computed from the temperature $T$ and Boltzmann constant $k_\text{B}$. Given the partition function, the free energy is defined as
\begin{equation}
\mathcal{F}(\vecR) \equiv -\beta^{-1} \log Z(\vecR).
\end{equation}
The CG force is the negative of the first derivative $\nabla_R \mathcal{F} \equiv -\mathbf{F}_{\text{CG}}$, while the CG Hessian is the second derivative $(\nabla_R \otimes \nabla_R) \mathcal{F} \equiv \mathbf{H}_{\text{CG}}$. To compute the latter, we need the former, which is
\begin{equation}
\nabla_R \mathcal{F} = -\frac{1}{\beta Z} \nabla_R Z = -\frac{1}{\beta Z}\int d\vecr\big[\nabla_R \delta(\boldsymbol{\Xi}_r \vecr - \vecR)\big]e^{-\beta \mathcal{H}(\vecr)},\label{force1Lin}
\end{equation}
However, we want to recast the derivative inside of the integral to be with respect to the AA coordinates $\vecr$. To do so, we use the identity
\begin{equation}
\nabla_R \delta(\boldsymbol{\Xi}_r \vecr - \vecR) = -\boldsymbol{\Xi}_F \left[\nabla_r \delta(\boldsymbol{\Xi}_r \vecr - \vecR)\right],\label{deltaDerivsLin}
\end{equation}
where $\boldsymbol{\Xi}_F$ is the following $3N \times 3n$ force-projection matrix:
\begin{equation}
\boldsymbol{\Xi}_F \equiv \left(\boldsymbol{\Xi}_r \boldsymbol{\Xi}_r^T\right)^{-1} \boldsymbol{\Xi}_r.
\end{equation}
For the C$_\alpha$ coarse-graining map, 
$\boldsymbol{\Xi}_r \boldsymbol{\Xi}_r^T = \mathbf{I}_{3N} \implies \boldsymbol{\Xi}_F = \boldsymbol{\Xi}_r$ ($\mathbf{I}_{3N}$ is the $3N \times 3N$ identity matrix), 
but we will continue to assume more general linear mappings here. Plugging \eqref{deltaDerivsLin} into \eqref{force1Lin} and integrating by parts yields
\begin{equation}
\nabla_R \mathcal{F}
= \frac{1}{Z}\int d\vecr\,\delta(\boldsymbol{\Xi}_r \vecr-\vecR) e^{-\beta \mathcal{H}(\boldsymbol{r})}\big(\boldsymbol{\Xi}_F \nabla_r \mathcal{H}\big).\label{force2Lin}
\end{equation}
In Hamiltonian mechanics, $-\nabla_r \mathcal{H}$ is identified as the AA force $\mathbf{F}_{\text{AA}}$, and so the expression for the first derivative $\nabla_R \mathcal{F}$ can be reinterpreted as an ensemble average conditioned on $\mathbf{R}$ (written as $\expval{\cdot}_R$):
\begin{equation}
\nabla_R \mathcal{F} = -\expval{\boldsymbol{\Xi}_F \mathbf{F}_{\text{AA}}}_R \implies \mathbf{F}_{\text{CG}} = \expval{\boldsymbol{\Xi}_F \mathbf{F}_{\text{AA}}}_R.\label{ensembleForce}
\end{equation}
As defined by the force matching framework, the CG force is the ensemble average of the projected AA forces. The linear identity here, and its nonlinear generalization in Appendix~\ref{app:genBlueMoon}, recover known mean-force results from the Blue Moon literature~\citep{Ciccotti2005, Kalligiannaki2015}.

The CG Hessian is computed by differentiating the ensemble average~\eqref{ensembleForce}. Through further path-integral manipulations (detailed in Appendix~\ref{app:genBlueMoon}), we obtain
\begin{equation}
\mathbf{H}_{\text{CG}} = -\langle{\boldsymbol{\Xi}_F\left(\nabla_r \mathbf{F}_{\text{AA}}\right) \boldsymbol{\Xi}_F^T}\rangle_R - \beta\boldsymbol{\Sigma}(\boldsymbol{\Xi}_F \mathbf{F}_{\text{AA}},\boldsymbol{\Xi}_F \mathbf{F}_{\text{AA}}),
\end{equation}
where $\boldsymbol{\Sigma}(\cdot,\cdot)$ is the covariance matrix. We further note that the AA Hessian $\mathbf{H}_{\text{AA}}$ is defined as the second derivative of the Hamiltonian, \textit{i.e.} $\mathbf{H}_{\text{AA}} \equiv (\nabla_r \otimes \nabla_r) \mathcal{H} = -\nabla_r \mathbf{F}_{\text{AA}}$. We then obtain our primary expression for the CG Hessian:
\begin{equation}\label{eq:cg_hessian_main}
\boxed{\mathbf{H}_{\text{CG}} = \langle{\boldsymbol{\Xi}_F \mathbf{H}_{\text{AA}} \boldsymbol{\Xi}_F^T}\rangle_R - \beta \boldsymbol{\Sigma}(\boldsymbol{\Xi}_F \mathbf{F}_{\text{AA}},\mathbf{\Xi}_F \mathbf{F}_{\text{AA}}).}
\end{equation}
The first term is the ensemble-averaged projected Hessian, which captures the average curvature of the AA potential energy surface as seen through the CG mapping. The second term is the force covariance correction, which captures the softening of the effective CG potential due to thermal fluctuations of the integrated-out AA degrees of freedom. At higher temperatures, the force covariance $\boldsymbol{\Sigma}$ grows faster than $\beta$ shrinks, so Term~2 contributes more to the effective CG curvature. 

These two terms carry distinct physical information that goes beyond what the force alone provides, while also containing fundamentally different computational properties. Term~1 depends only on the AA potential and the CG mapping, making it model-independent, while Term~2 depends on the CG model's current force predictions through the force residual. We exploit this decomposition in the methodology that follows.

\section{Methodology}
\label{sec:methodology}

We now describe how to turn the CG Hessian identity \eqref{eq:cg_hessian_main} into a practical training objective via computationally feasible Hessian-vector product (HVP) calculations. We formulate the stochastic HVP matching loss (Section~\ref{sec:hvp_method}) and present the training pipeline (Section~\ref{sec:pipeline}).

\subsection{Stochastic Hessian matching via Hessian-vector products}
\label{sec:hvp_method}

\paragraph{Intractability of full Hessian matching.}
For a CG system with $d = 3N$ degrees of freedom, the full Hessian $\mathbf{H}_{\text{CG}} \in \mathbb{R}^{d \times d}$ requires $\mathcal{O}(d^2)$ storage and $\mathcal{O}(d)$ force evaluations to fully construct. For the chignolin protein ($d = 30$), this is feasible though slow, but for larger proteins $d$ can scale to the thousands, which makes explicit Hessian construction intractable.

\paragraph{The HVP reformulation.}
Instead of matching the entire Hessian, we match its action on $K$ random probe vectors, each providing curvature information along one particular direction of the energy landscape. For each training frame, we sample unit vectors $\{\mathbf{v}_k\}_{k=1}^K$ by drawing $\mathbf{z}_k \sim \mathcal{N}(0, \mathbf{I}_d)$ and normalizing to unit length. Multiplying both sides of \eqref{eq:cg_hessian_main} by $\mathbf{v}_k$ gives the target HVP,
\begin{equation}
\label{eq:hvp_target}
\mathbf{H}_\text{CG}\mathbf{v}_k 
= \underbrace{\boldsymbol{\Xi}_F\left(\mathbf{H}_\text{AA}
\boldsymbol{\Xi}_F^T \mathbf{v}_k\right)}_{\text{Term~1}} 
- \underbrace{\beta\, \delta\mathbf{J}\left(\delta\mathbf{J}^T 
\mathbf{v}_k\right)}_{\text{Term~2}}.
\end{equation}
These two terms play distinct physical roles in the Hessian-matching objective. For Term~1, which is the projected AA HVP, we embed the CG probe $\mathbf{v}_k$ into AA space via $\boldsymbol{\Xi}_F^T$, acting only on the C$_{\alpha}$ atoms. The atomistic Hessian $\mathbf{H}_\text{AA}$ then returns the AA curvature along this restricted direction, and $\boldsymbol{\Xi}_F$ projects this curvature back to CG space. Term~1 depends only on the AA potential and the CG mapping, so it is model-independent and can therefore be precomputed once before training.

Term~2 is the covariance correction, driven by the force residual,
\begin{equation}
\label{eq:force_residual}
\delta\mathbf{J} \equiv \boldsymbol{\Xi}_F \mathbf{F}_\text{AA} 
- \mathbf{F}_\text{NN}.
\end{equation}
This measures how far the CG model's current mean-force prediction $\mathbf{F}_\text{NN}$ is from the projected AA force $\boldsymbol{\Xi}_F \mathbf{F}_\text{AA}$. Term~2 is model-dependent because the residual changes with every training iteration, but since the forces are already computed during the forward pass, the term can be computed on-the-fly at negligible additional cost.

Each HVP is a vector of size $d$, and we compute only $K$ of them per frame. This gives a total cost of $\mathcal{O}(Kd)$. Thus, this reformulation lets us instill second-order curvature information into the CG potential without ever constructing the full $d \times d$ Hessian matrix.

\paragraph{Loss function and unbiased estimation.}
Given a set of $T$ training frames, the standard force matching loss is the mean-squared error between the model's predicted forces and the AA forces projected to CG space~\citep{Izvekov2005,Noid2008},
\begin{equation}
\label{eq:force_matching}
\mathcal{L}_\text{FM} = \frac{1}{T} \sum_{t=1}^{T} \frac{1}{3 N^{(t)}} 
\big\| \mathbf{F}_\text{NN}^{(t)} - 
\boldsymbol{\Xi}_F \mathbf{F}_\text{AA}^{(t)} \big\|^2,
\end{equation}
where $^{(t)}$ designates quantities at frame $t$ and $\boldsymbol{\Xi}_F$ is the force-projection matrix. We supplement this with the following HVP-matching loss:
\begin{equation}
\label{eq:hvp_loss}
\mathcal{L}_\text{HVP} = \frac{1}{T}\sum_{t=1}^{T} \frac{1}{K}\sum_{k=1}^{K} 
\frac{1}{3 N^{(t)}} \big\| 
(\mathbf{H}_\text{NN}\mathbf{v}_k)^{(t)} 
- (\mathbf{H}_\text{CG}\mathbf{v}_k)^{(t)} \big\|^2,
\end{equation}
where $(\mathbf{H}_\text{NN}\mathbf{v}_k)^{(t)}$ is the CG model's prediction for the HVP. A set of $K$ probe vectors $\{\mathbf{v}_k^{(t)}\}$ is chosen independently for each frame $t$, though seeded deterministically by the frame index for reproducibility. For probe vectors uniformly distributed on the unit sphere, $\mathbb{E}_{\mathbf{v}}[\|\mathbf{H}\mathbf{v}\|^2] = \|\mathbf{H}\|_F^2 / d$ for any matrix $\mathbf{H}$, where $||\cdot||_F$ is the Frobenius norm. So, each per-frame term is an unbiased stochastic estimate of the full Hessian-matching objective. The total training objective combines force and HVP matching,
\begin{equation}
\label{eq:total_loss}
\mathcal{L} = w_\text{FM}\,\mathcal{L}_\text{FM} + w_\text{HVP}\,
\mathcal{L}_\text{HVP}.
\end{equation}

\subsection{Training pipeline}
\label{sec:pipeline}

The overall pipeline is shown schematically in Figure~\ref{fig:pipeline}. We summarize each step here.

\paragraph{Probe selection.}
For each frame $t = 1, \dots, T$, we generate $K$ probes using a deterministic seeding scheme $s^{(t)} = t$. Probe consistency for HVP matching is guaranteed 
by this scheme, in which the same seed produces identical probes per frame on any device and across multi-GPU setups. Term~1 targets can therefore be precomputed once and reused across any subsequent model trained with the HVP loss. 

\paragraph{Precomputing Term~1 target.}
Since the Term~1 targets in \eqref{eq:hvp_target} are model-independent, we compute them once before training and store the results. We use the Amber14 force field~\citep{Maier2015} evaluated in OpenMM~\citep{Eastman2017} with no nonbonded cutoff. For each frame, the CG probe is embedded into AA space via $\tilde{\mathbf{v}}_k \equiv \boldsymbol{\Xi}_F^T \mathbf{v}_k$, and the AA HVP is approximated via central finite differences,
\begin{equation}
\label{eq:fd_hvp}
\mathbf{H}_\text{AA}\tilde{\mathbf{v}}_k \approx 
-\frac{\mathbf{F}_\text{AA}(\mathbf{r} + \varepsilon\tilde{\mathbf{v}}_k) - 
\mathbf{F}_\text{AA}(\mathbf{r} - \varepsilon\tilde{\mathbf{v}}_k)}{2\varepsilon},
\end{equation}
where $\varepsilon$ is some small finite-difference step. The result is projected back to CG space via 
$\boldsymbol{\Xi}_F$.

\paragraph{Computing the CG model's HVP.}
The CG-side HVP $\mathbf{H}_\text{NN}\,\mathbf{v}_k$ is obtained via two sequential applications of automatic differentiation, which is the primary source of computational overhead in our pipeline. First, the energy $W_\text{NN}(\mathbf{R})$ is differentiated with respect to positions to produce the forces $\mathbf{F}_\text{NN} = -\nabla_R W_\text{NN}$, with the computation graph retained. Note that the forces are not an explicit output of the GNN, so this step is what creates them as a differentiable node in the graph in the first place. Subsequently, the inner product $\mathbf{F}_\text{NN} \cdot \mathbf{v}_k$ is differentiated with respect to positions, yielding the HVP:
\begin{equation}
\mathbf{H}_\text{NN}\,\mathbf{v}_k = -\nabla_R(\mathbf{F}_\text{NN} 
\cdot \mathbf{v}_k).
\end{equation}
The graph from the first differentiation step is reused across all $K$ probes, so the marginal cost of each additional probe is a single backward pass through the network.

\paragraph{Computing Term~2 correction.}
The covariance correction $\beta\,\delta\mathbf{J}(\delta\mathbf{J}^T\mathbf{v}_k)$ is computed from the force residual $\delta\mathbf{J}$, so it can only be computed online during training. However, because this quantity is built from forces already produced during the forward pass ($\mathbf{F}_\text{NN}$ and $\boldsymbol{\Xi}_F \mathbf{F}_\text{AA}$), it adds negligible computational overhead. The residual is detached from the computation graph, so gradients do not flow through the target. Term~2 is then subtracted from the precomputed Term~1 target to form the full HVP target.

\paragraph{Training loop.}
At each iteration, the GNN forward pass produces the energy, forces, and HVPs via the two-pass autograd pipeline. Probe vectors are regenerated at training time, avoiding the memory overhead of saving $K$ vectors per frame. Our pipeline supports two variants of $\mathcal{L}_\text{HVP}$ with the covariance correction either enabled or disabled. If the covariance correction is included, then Term~2 is assembled on GPU from the force residual and subtracted from Term~1 before the loss computation. The combined loss $\mathcal{L}$ is then backpropagated through the network, with no gradients flowing through the HVP target since it is either precomputed or detached.

\begin{figure}[t]
\centering
\resizebox{0.92\textwidth}{!}{%
\begin{tikzpicture}[
    >=Stealth,
    every node/.style={font=\small},
    box/.style={
        rectangle, rounded corners=3pt, 
        minimum width=3.4cm, minimum height=1.1cm,
        draw=#1!80!black, fill=#1!12, line width=0.5pt,
        text=black, align=center
    },
    smallbox/.style={
        rectangle, rounded corners=3pt, 
        minimum width=2.6cm, minimum height=1.1cm,
        draw=#1!80!black, fill=#1!12, line width=0.5pt,
        text=black, align=center
    },
    widebox/.style={
        rectangle, rounded corners=3pt,
        minimum width=7.0cm, minimum height=1.15cm,
        draw=#1!80!black, fill=#1!12, line width=0.5pt,
        text=black, align=center
    },
    container/.style={
        rectangle, rounded corners=6pt,
        draw=black, line width=0.8pt, dashed,
        inner sep=6pt
    },
    arr/.style={->, >=Stealth, line width=0.6pt, color=black},
    dasharr/.style={->, >=Stealth, line width=0.6pt, color=black, dashed},
    annot/.style={text=black},
]
 
\definecolor{probecolor}{HTML}{8B7FC7}
\definecolor{cgcolor}{HTML}{2A9D8F}
\definecolor{aacolor}{HTML}{D4952B}
\definecolor{t2color}{HTML}{C75480}
\definecolor{losscolor}{HTML}{C85A3A}
 
\node[box=probecolor] (probes) at (0,0.5) {
    \underline{\textbf{Shared probes}~$\mathbf{v}_1, \ldots, \mathbf{v}_K$}\\[3.5pt]
    {\footnotesize Deterministic seed $s^{(t)}=t$}
};
 
\node[box=cgcolor] (gnn) at (-5.2, -1.9) {
    \textbf{\underline{Energy function}}\\[3.5pt]
    \textbf{GNN} $W_\text{NN}(\mathbf{R})$
};
 
\node[box=cgcolor] (hvpnn) at (-5.2, -3.5) {
    \textbf{\underline{Autograd HVP}}\\[3.5pt]
    {\footnotesize $\mathbf{F}_\text{NN} = -\nabla_R W_{\text{NN}}$}\\
    {\footnotesize $\mathbf{H}_\text{NN}\mathbf{v}_k = -\nabla_R\big(\mathbf{F}_{\text{NN}} \mathbf{v}_k\big)$}
};
 
\begin{scope}[on background layer]
\node[container, fit=(gnn)(hvpnn)] (cgbox) {};
\end{scope}
 
\node[box=aacolor] (fd) at (5.2, -2.0) {
    \textbf{\underline{Finite differences}}\\[3.5pt]
    {\footnotesize $\mathbf{H}_{\text{AA}}\mathbf{v}_k = -\dfrac{\mathbf{F}(\mathbf{r}+\varepsilon\tilde{\mathbf{v}})-\mathbf{F}(\mathbf{r}-\varepsilon\tilde{\mathbf{v}})}{2\varepsilon}$}
};
 
\node[box=aacolor] (term1) at (5.2, -3.6) {
    \textbf{\underline{Term 1 target}}\\[3.5pt]
    {\footnotesize $\boldsymbol{\Xi}_{F}\big(\mathbf{H}_\text{AA}\boldsymbol{\Xi}_{F}^T\mathbf{v}_k\big)$}
};
 
\begin{scope}[on background layer]
\node[container, fit=(fd)(term1)] (aabox) {};
\end{scope}
 
\node[smallbox=t2color] (term2) at (0, -2.8) {
    \textbf{\underline{Term 2}}~{\footnotesize(online)}\\[3.5pt]
    {\footnotesize $\beta \, \delta{\mathbf{J}}\big(\delta{\mathbf{J}}^T\mathbf{v}_k\big)$}
};
 
\node[widebox=losscolor] (loss) at (0, -6) {
    \textbf{\underline{Combined loss}}\\[3.5pt]
    {\footnotesize $\mathcal{L} \;=\; w_\text{FM}\,\mathcal{L}_\text{FM} \;+\; w_\text{HVP}\,\mathcal{L}_\text{HVP}$}
};
 
 
\draw[arr] (probes.south west) ++(0.3,0) -- (cgbox.north -| gnn) node[midway, above, sloped] {CG space};
\draw[arr] (probes.south east) ++(-0.3,0) -- (aabox.north -| fd) node[midway, above, sloped] {\makecell{AA space\\{\scriptsize{(precomputed)}}}};
 
\draw[arr] (gnn) -- (hvpnn);
\draw[arr] (fd) -- (term1);
 
\draw[dasharr] (cgbox.east |- term2) -- (term2.west);
 
\draw[dasharr] (aabox.west |- term2) -- (term2.east);
 
\draw[arr] (cgbox.south -| hvpnn) -- ([xshift=0.3cm]loss.north west) node[midway, below, sloped] {prediction};
 
\draw[arr] (term2.south) -- (loss.north) node[midway, above, sloped] {target};

\end{tikzpicture}%
}
\caption{%
Overview of the HVP matching pipeline. 
Shared probe vectors $\mathbf{v}_k$ are generated from deterministic per-frame seeds, and are used on both sides of the matching. 
\textbf{Left:}~the GNN produces energy and forces, and two-pass automatic differentiation yields $H_\text{NN}\mathbf{v}_k$ (the prediction). 
\textbf{Right:}~central finite differences on the AA force field produce Term~1 targets, precomputed in one shot before training. 
\textbf{Center:}~the force residual $\delta{\mathbf{J}}$ is computed from the CG model's force prediction and the projected AA forces in order to evaluate Term~2 online. 
The combined loss matches both forces ($\mathcal{L}_\text{FM}$) and curvature ($\mathcal{L}_\text{HVP}$) factors.%
}
\label{fig:pipeline}
\end{figure}

\section{Experiments}
\label{sec:experiments}

To isolate the effect of curvature supervision, we trained three different model variants with different losses, described below. We both trained and evaluated on the same benchmark suite of~\citet{Aghili2025}, which consists of 9 fast-folding proteins simulated in explicit solvent, spanning a range of sizes from chignolin (10 CG beads) to $\lambda$-repressor (80 CG beads); this constitutes an in-distribution experiment. The out-of-distribution case, which is less controlled, is considered separately in Appendix~\ref{app:generalization}.

\paragraph{Model variants.}
We compared three training configurations with identical architecture and hyperparameters:
\begin{itemize}
\item \textbf{FM}: force matching only;
\item \textbf{FM+AA$_\text{p}$}: force and HVP matching with Term~1 (the AA Hessian projection) with precomputed targets; and
\item \textbf{FM+AA$_\text{p}$+Cov}: the full objective including the Term~2 covariance correction computed online during training.
\end{itemize}
The loss weights in \eqref{eq:total_loss} were set to $w_{\text{FM}} = 1$ and either $w_\text{HVP} = 0$ for FM or $w_\text{HVP} = 0.01$ for FM+AA$_\text{p}$ and FM+AA$_\text{p}$+Cov.

To represent the energy function, all models use the SchNet-based graph neural network from TorchMD-Net~\citep{Doerr2021,pelaez2024torchmdnet}, following the CGSchNet framework~\citep{Husic2020} for coarse-grained protein potentials. The architecture comprises 4 message-passing layers, 64-dimensional embeddings, 12 radial basis functions (exponential normal), and SiLU activations, with a bead-bead interaction cutoff spanning 3.0--12.0~\AA.

All three models were trained on the nine benchmark proteins until convergence, plateauing at approximately 20 epochs. Training used AdamW~\citep{Loshchilov2019} at a learning rate of $10^{-4}$ with batch size 200 on 4$\times$A40 GPUs. All simulations and training data corresponded to $T=300$\,K, giving the Term~2 prefactor $\beta = 1/(k_\text{B} T) = 1.677$~mol/kcal. The HVPs were computed with $K = 8$ probe vectors per frame, with $\varepsilon = 10^{-5}$~nm for the finite-differences evaluation~\eqref{eq:fd_hvp}.

All three variants trained at comparable per-epoch cost. On a compute-bound single-chain run, the HVP objective added $\sim$16\% wall-clock overhead over force matching, but on the benchmark models this overhead was negligible relative to data loading and variability on the shared cluster. The covariance correction term adds negligible per-step cost and requires no architectural changes, while the HVP Term~1 targets require a one-time precomputation and are reused across all HVP-matching runs.

\paragraph{Evaluation.}
Following the benchmarking methodology of \citet{Aghili2025}, we ran CG molecular dynamics from each trained potential and compared trajectory metric distributions against AA reference simulations. We consider two classes of benchmark metrics. The first come from time-lagged independent component analysis (TICA)~\citep{Perez2013} of the free energy, which identifies and ranks the directions of slowest collective motion in molecular trajectories. In particular, TIC0 captures the slowest (typically folding/unfolding) transition and TIC1 the second slowest. We project our trajectories onto these components and compute distributional metrics on the projections. The second class measures the error in a model's learned molecular structure.


\begin{table}[ht]
\centering
\setlength{\tabcolsep}{6pt}
\begin{tabular}{@{}l@{\hspace{6pt}}lrrrrr@{}}
\toprule
 & & \multicolumn{2}{c}{\textbf{Slow-mode (KL)}} & \multicolumn{3}{c}{\textbf{Structural (W1)}} \\
\cmidrule(lr){3-4} \cmidrule(lr){5-7}
Protein & Model & \makecell{TIC0 $\downarrow$} & \makecell{TIC1 $\downarrow$} & \makecell{Length $\downarrow$\\$(\times 10^{-2})$} & \makecell{Angle $\downarrow$} & \makecell{RoG $\downarrow$} \\
\midrule
\multirow{3}{*}{\shortstack[l]{Chignolin\\ $N=10$}}
 & FM               & $2.23_{\pm 0.53}$ & $1.64_{\pm 0.53}$ & $0.11_{\pm 0.03}$ & $0.080_{\pm 0.010}$ & $0.171_{\pm 0.005}$ \\
 & +AA$_{\text{p}}$ & $\mathbf{0.40_{\pm 0.11}}$ & $\mathbf{0.64_{\pm 0.13}}$ & $\mathbf{0.03_{\pm 0.01}}$ & $0.060_{\pm 0.009}$ & $0.132_{\pm 0.009}$ \\
 & +Cov             & $1.23_{\pm 0.24}$ & $0.78_{\pm 0.05}$ & $\mathbf{0.03_{\pm 0.01}}$ & $\mathbf{0.033_{\pm 0.005}}$ & $\mathbf{0.120_{\pm 0.002}}$ \\
\midrule
\multirow{3}{*}{\shortstack[l]{Trp-cage\\ $N=20$}}
 & FM               & $0.94_{\pm 0.39}$ & $1.04_{\pm 0.45}$ & $0.83_{\pm 0.17}$ & $0.096_{\pm 0.020}$ & $\mathbf{0.107_{\pm 0.025}}$ \\
 & +AA$_{\text{p}}$ & $1.05_{\pm 0.05}$ & $1.18_{\pm 0.28}$ & $0.04_{\pm 0.02}$ & $\mathbf{0.063_{\pm 0.011}}$ & $0.108_{\pm 0.007}$ \\
 & +Cov             & $\mathbf{0.78_{\pm 0.10}}$ & $\mathbf{0.98_{\pm 0.06}}$ & $\mathbf{0.04_{\pm 0.01}}$ & $0.065_{\pm 0.008}$ & $0.134_{\pm 0.009}$ \\
\midrule
\multirow{3}{*}{\shortstack[l]{BBA\\ $N=28$}}
 & FM               & $\mathbf{2.90_{\pm 1.08}}$ & $\mathbf{0.60_{\pm 0.37}}$ & $1.96_{\pm 1.55}$ & $0.056_{\pm 0.010}$ & $0.325_{\pm 0.237}$ \\
 & +AA$_{\text{p}}$ & $4.78_{\pm 0.76}$ & $0.72_{\pm 0.40}$ & $0.29_{\pm 0.16}$ & $\mathbf{0.051_{\pm 0.011}}$ & $0.254_{\pm 0.031}$ \\
 & +Cov             & $3.91_{\pm 0.32}$ & $0.82_{\pm 0.26}$ & $\mathbf{0.07_{\pm 0.02}}$ & $0.081_{\pm 0.014}$ & $\mathbf{0.236_{\pm 0.034}}$ \\
\midrule
\multirow{3}{*}{\shortstack[l]{WW domain\\ $N=34$}}
 & FM               & $2.05_{\pm 1.06}$ & $1.41_{\pm 0.14}$ & $\mathbf{0.13_{\pm 0.06}}$ & $\mathbf{0.075_{\pm 0.008}}$ & $0.199_{\pm 0.066}$ \\
 & +AA$_{\text{p}}$ & $2.92_{\pm 1.87}$ & $1.65_{\pm 0.36}$ & $0.64_{\pm 0.09}$ & $0.084_{\pm 0.004}$ & $0.259_{\pm 0.018}$ \\
 & +Cov             & $\mathbf{1.66_{\pm 0.43}}$ & $\mathbf{1.30_{\pm 0.19}}$ & $0.31_{\pm 0.14}$ & $0.096_{\pm 0.010}$ & $\mathbf{0.195_{\pm 0.011}}$ \\
\midrule
\multirow{3}{*}{\shortstack[l]{Protein B\\ $N=47$}}
 & FM               & $3.50_{\pm 2.19}$ & $4.74_{\pm 3.21}$ & $0.19_{\pm 0.31}$ & $0.153_{\pm 0.051}$ & $0.167_{\pm 0.035}$ \\
 & +AA$_{\text{p}}$ & $\mathbf{1.77_{\pm 1.14}}$ & $\mathbf{0.60_{\pm 0.34}}$ & $0.94_{\pm 1.08}$ & $\mathbf{0.072_{\pm 0.015}}$ & $0.197_{\pm 0.014}$ \\
 & +Cov             & $1.96_{\pm 0.39}$ & $0.81_{\pm 0.41}$ & $\mathbf{0.04_{\pm 0.01}}$ & $0.134_{\pm 0.019}$ & $\mathbf{0.122_{\pm 0.015}}$ \\
\midrule
\multirow{3}{*}{\shortstack[l]{Homeodomain\\ $N=54$}}
 & FM               & $4.50_{\pm 1.97}$ & $3.93_{\pm 0.81}$ & $0.04_{\pm 0.00}$ & $0.045_{\pm 0.008}$ & $\mathbf{0.069_{\pm 0.025}}$ \\
 & +AA$_{\text{p}}$ & $\mathbf{2.33_{\pm 0.50}}$ & $\mathbf{2.24_{\pm 0.90}}$ & $0.06_{\pm 0.02}$ & $0.044_{\pm 0.004}$ & $0.106_{\pm 0.019}$ \\
 & +Cov             & $2.50_{\pm 1.16}$ & $3.29_{\pm 1.97}$ & $\mathbf{0.03_{\pm 0.01}}$ & $\mathbf{0.040_{\pm 0.002}}$ & $0.098_{\pm 0.005}$ \\
\midrule
\multirow{3}{*}{\shortstack[l]{Protein G\\ $N=56$}}
 & FM               & $3.62_{\pm 0.65}$ & $1.93_{\pm 0.30}$ & $0.20_{\pm 0.07}$ & $0.075_{\pm 0.016}$ & $\mathbf{0.245_{\pm 0.022}}$ \\
 & +AA$_{\text{p}}$ & $3.35_{\pm 0.81}$ & $2.20_{\pm 0.29}$ & $0.60_{\pm 0.11}$ & $0.089_{\pm 0.015}$ & $0.270_{\pm 0.013}$ \\
 & +Cov             & $\mathbf{2.74_{\pm 1.53}}$ & $\mathbf{1.56_{\pm 0.71}}$ & $\mathbf{0.17_{\pm 0.02}}$ & $\mathbf{0.062_{\pm 0.004}}$ & $0.254_{\pm 0.025}$ \\
\midrule
\multirow{3}{*}{\shortstack[l]{$\alpha$3D\\ $N=73$}}
 & FM               & $1.89_{\pm 1.99}$ & $2.77_{\pm 0.75}$ & $0.73_{\pm 1.49}$ & $0.043_{\pm 0.003}$ & $0.141_{\pm 0.030}$ \\
 & +AA$_{\text{p}}$ & $1.75_{\pm 1.29}$ & $1.71_{\pm 0.79}$ & $\mathbf{0.06_{\pm 0.01}}$ & $0.046_{\pm 0.005}$ & $\mathbf{0.122_{\pm 0.013}}$ \\
 & +Cov             & $\mathbf{0.67_{\pm 0.16}}$ & $\mathbf{0.90_{\pm 0.24}}$ & $0.09_{\pm 0.02}$ & $\mathbf{0.040_{\pm 0.004}}$ & $0.161_{\pm 0.024}$ \\
\midrule
\multirow{3}{*}{\shortstack[l]{$\lambda$-repressor\\ $N=80$}}
 & FM               & $5.83_{\pm 1.96}$ & $4.42_{\pm 3.36}$ & $0.17_{\pm 0.05}$ & $0.066_{\pm 0.034}$ & $0.289_{\pm 0.084}$ \\
 & +AA$_{\text{p}}$ & $\mathbf{3.84_{\pm 2.42}}$ & $3.79_{\pm 2.03}$ & $0.18_{\pm 0.09}$ & $0.064_{\pm 0.013}$ & $0.240_{\pm 0.106}$ \\
 & +Cov             & $5.88_{\pm 1.04}$ & $\mathbf{2.85_{\pm 0.92}}$ & $\mathbf{0.11_{\pm 0.01}}$ & $\mathbf{0.044_{\pm 0.002}}$ & $\mathbf{0.217_{\pm 0.030}}$ \\
\bottomrule
\end{tabular}
\caption{In-distribution benchmark, consisting of both slow-mode metrics (KL divergences over TIC0 and TIC1 projections) and structural statistics (W1 on bond-length, bond-angle, and radius-of-gyration distributions). Metrics are computed over 5 seeds (mean$_{\pm\text{std}}$), with best per protein in bold; lower mean followed by lower spread is better.}
\label{tab:id_combined}
\end{table}

All metrics are computed over 20 independent CG MD replica trajectories per protein, averaging out variance from MD sampling. Each model variant was trained from a single random seed, and reported metrics are means over five independent MD sampling seeds. We evaluate our models using the following metrics:
\begin{itemize}
\item Kullback--Leibler (KL) divergence on TIC0 and TIC1 projections; and
\item Wasserstein-1 (W1) distance on bond-length, bond-angle, and  radius-of-gyration (RoG) distributions.
\end{itemize} 
Table~\ref{tab:id_combined} reports both slow-mode and structural accuracy for all three variants across the benchmark set (see Appendix~\ref{app:inDistSupplement} for select plots of learned densities and additional metrics). Supplementing force matching with HVP matching improves slow-mode accuracy on eight of nine proteins based on KL divergences along TIC0 and TIC1, including the largest.

Which HVP variant performs best is protein-dependent. On the smallest protein, chignolin, the projected-Hessian term alone (FM+AA$_\text{p}$) is best, reducing TIC0~KL by 82\% and TIC1~KL by 61\% relative to force matching, while the covariance correction (FM+AA$_\text{p}$+Cov) more often produces the best slow-mode accuracy on the mid-sized and larger proteins. One possible explanation is that the force residual $\delta\mathbf{J}$ driving Term~2 captures more physically meaningful thermal fluctuations in larger, slower-converging systems, though our experiments do not isolate this effect.

The gains are notable on the two largest proteins. On $\lambda$-repressor, HVP matching improves both slow modes relative to force matching, with FM+AA$_\text{p}$ reducing TIC0~KL by 34\% and FM+AA$_\text{p}$+Cov reducing TIC1~KL by 36\%. On $\alpha$3D, the covariance correction reduces TIC0~KL by 65\% and TIC1~KL by 68\%, while the projected-Hessian term alone yields more marginal slow-mode benefit. The fact that Term~1 alone can be insufficient while the full two-term objective restores strong performance suggests that the model-dependent covariance correction provides an important component of the curvature signal for some proteins, consistent with its role in the CG Hessian identity.

Beyond the slow modes, HVP matching also improves local structural fidelity, especially on the bond-length distributions. An HVP-matching variant achieves the lowest bond-length Wasserstein distance on eight of nine proteins, often by a large margin. For example, Trp-cage improves by 95\%, BBA by 96\%, and $\alpha$3D by 92\%. This is consistent with the Hessian directly governing the stiffness of bonded interactions. Improvements on bond angles and radius of gyration are more mixed and protein-dependent.

\section{Discussion}

We have presented a framework for instilling second-order curvature information into coarse-grained neural potentials via stochastic HVP matching. Our key theoretical result is the decomposition of the CG Hessian into a model-independent projected AA Hessian (Term~1) and a model-dependent covariance correction (Term~2). Term~1 targets are precomputed once and reused across training runs, while Term~2 is assembled from quantities already available during the forward pass, enabling the addition of curvature supervision to any force matching pipeline with no architectural changes and $\mathcal{O}(Kd)$ additional cost per frame. We note that this is limited to C$_\alpha$ linear coarse-graining of single-chain proteins, leaving the extension to diverse multi-chain complexes, non-standard CG mappings, and the nonlinear Hessian identity in Appendix~\ref{app:genBlueMoon} as natural next steps.

Our ablation experiments on nine fast-folding benchmark proteins demonstrate a clear empirical benefit from curvature supervision. In the in-distribution setting, some form of HVP matching outperforms plain force matching on both TIC0 and TIC1 for eight of nine proteins while also improving the bond-length distribution on eight of nine proteins. The relative performance of the two HVP variants is protein-dependent rather than determined by system size, with the covariance correction providing the larger benefit in several instances. In the case of $\alpha$3D, the projected-Hessian term alone provides little improvement over force matching (though it does reduce TIC~1), whereas the full two-term objective substantially reduces KL divergence along both TIC0 and TIC1, suggesting that the model-dependent covariance correction can provide an important component of the curvature signal in some systems. These conclusions are further bolstered by supplementary slow-mode and structural metrics, which are shown in Appendix~\ref{app:inDistSupplement}.

Several limitations point toward avenues for future work. Probe coverage may bottleneck the largest systems, since $K = 8$ probes sample a diminishing fraction of the $d$-dimensional curvature space as protein size grows. Adaptive probe selection along physically important directions could improve efficiency without increasing $K$. Furthermore, the loss weight $w_\text{HVP} = 0.01$ was chosen heuristically, and Term~1 and Term~2 are weighted equally in the target \eqref{eq:hvp_target}. A systematic study of $w_\text{HVP}$, alongside an interpolation coefficient $\alpha \in [0,1]$ that continuously scales the covariance correction, could potentially replace the binary FM+AA$_\text{p}$ vs. FM+AA$_\text{p}$+Cov ablation with a single hyperparameter whose optimal value may correlate with the system's attributes.

Another important direction of this methodology is its extension to generalization. In a preliminary out-of-distribution experiment (Appendix~\ref{app:generalization}), we find that the in-distribution improvements over force matching also carry over to proteins held out entirely from training, though the results are generally more mixed. While a full study of transferability is beyond the scope of this work, these results suggest that curvature supervision is a promising direction for improving the generalization of CG potentials.

Ultimately, our results indicate that the accuracy of CG neural potentials can be improved not by increasing model capacity or data scale, but by enriching the physical content of the training signal. The same GNN architecture, when trained with higher-order physical targets, more faithfully reproduces the slow collective dynamics of the underlying system without changes to the model architecture or training data and at modest additional cost. More broadly, when we train a neural network to reproduce not only forces but also the shape of an energy landscape, higher-order derivatives of the target potential provide information that bolsters modeling beyond what first-order matching alone achieves.
\newpage

\subsection*{AI use statement}
In this work, we used generative AI tools for certain research and writing subtasks. For research execution, AI assistance was used to help implement portions of the training script, Hessian-vector-product precomputation, results visualization mechanisms, and job-launching scripts. For literature discovery, we used AI-assisted deep search to identify and verify potentially relevant related work. For the preparation of the manuscript, AI assistance was used to format tables and figures, to help revise the wording of some passages and check and format bibliography entries, and to cross-check reported numerical results against our experimental outputs for accuracy and consistency. The theoretical derivations, namely the CG Hessian identity in Section~\ref{sec:derivation} and its nonlinear generalization and symmetry proof in Appendix~\ref{app:genBlueMoon}, were derived by the authors without AI assistance.

We have reviewed all AI-assisted work. All AI-assisted code was verified and tested by the authors. All of the experimental results reported in the paper were produced by our existing experimental pipeline and also independently cross-checked against experimental outputs. AI-suggested related works were verified through manual review of the given sources. We take responsibility for the final content of this work including the text, claims, results, and any other artifacts produced with the assistance of generative AI.

\subsection*{Ethics statement}

This work develops a training method for coarse-grained molecular dynamics potentials and is evaluated entirely on publicly available simulation data of small proteins. It does not involve human subjects, personally identifiable information, or sensitive data, and we do not foresee direct harmful applications. Coarse-grained molecular simulation is a broadly used tool in the physical and life sciences, and by improving the accuracy of learned CG potentials at low computational cost, this work is intended to support basic scientific research such as the study of protein dynamics.

\subsection*{Reproducibility statement}

The CG Hessian identity underlying our method is derived in Section~\ref{sec:derivation}, with the general nonlinear case in Appendix~\ref{app:genBlueMoon}. The stochastic HVP-matching objective is defined in Section~\ref{sec:hvp_method}, and Section~\ref{sec:pipeline} details the training pipeline, including the deterministic per-frame probe seeding, the finite-difference precomputation of the Term~1 targets, and the online covariance correction. Section~\ref{sec:experiments} specifies the model architecture, loss weights, number of probes, finite-difference step, optimizer, learning rate, batch size, and hardware. In-distribution models are trained and evaluated on the benchmark of \citet{Aghili2025}, whose simulation and evaluation protocol we follow. All metrics are reported as mean $\pm$ standard deviation over five independent MD sampling seeds. Complete per-protein results, together with training and validation loss curves, are given in Appendix~\ref{app:inDistSupplement} (in-distribution) and Appendix~\ref{app:generalization} (out-of-distribution, including the full list of training proteins).

\subsubsection*{Acknowledgments}
This work was partly supported by National Science Foundation (NSF) STTR grant 2451680 ``Accelerating Molecular Dynamics using Coarse-Grained Neural Network Potentials.'' W.G.N.\ acknowledges support from the NSF under Grant Nos.\ CHE-2154433 and CHE-2502094.

\newpage
\bibliography{iclr2027_conference}
\bibliographystyle{iclr2027_conference}
\newpage
\appendix
\section{The Hessian for nonlinear coarse graining}\label{app:genBlueMoon}

In Section \ref{sec:derivation} of the main paper, we derived the Hessian for linear coarse-graining maps. The first-derivative results recover the local mean force expression of \citet{Kalligiannaki2015}, while our novel contribution is the Hessian identity in \eqref{hessianGen}. Here, we find the more general expression. We again use the Blue Moon ensemble and assume $n$ atoms and $N < n$ beads in three dimensions. This time, the partition function is
\begin{equation}
Z(\vecR) = \int d\vecr\,\delta(\boldsymbol{\xi}_r(\vecr) - \vecR) e^{-\beta \mathcal{H}(\vecr)},\ \ \ \ \beta \equiv \frac{1}{k_{\text{B}}T}.
\end{equation}
Again, $\vecr \in \mathbb{R}^{3n}$ and $\vecR \in \mathbb{R}^{3N}$ are elements of the AA space and the CG space, respectively. $\boldsymbol{\xi}_r: \mathbb{R}^{3n} \to \mathbb{R}^{3N}$ is the coarse-graining function.

\subsection{Computing the coarse-grained force}

The computation of the first derivative of the free energy $\mathcal{F}$ is very similar to the linear case. The starting point is
\begin{equation}
\nabla_R \mathcal{F} = -\frac{1}{\beta Z} \nabla_R Z = -\frac{1}{\beta Z}\int d\vecr\big[\nabla_R \delta(\boldsymbol{\xi}_r(\vecr) - \vecR)\big]e^{-\beta \mathcal{H}(\vecr)},\label{force1Gen}
\end{equation}
and we again need to rewrite the derivative to be in terms of $\vecr$. For the general case, the necessary formula is
\begin{equation}
\nabla_R \delta(\boldsymbol{\xi}_r(\vecr) - \vecR) = -\boldsymbol{\Xi}_F \left[\nabla_r \delta(\boldsymbol{\xi}_r(\vecr) - \vecR)\right],\label{deltaDerivsGen}
\end{equation}
where this time we define the $3N \times 3n$ matrix $\boldsymbol{\Xi}_F$ in terms of the $3N \times 3n$ Jacobian of $\boldsymbol{\xi}_r$:
\begin{equation}
\boldsymbol{\Xi}_F \equiv \left(\boldsymbol{J}_\xi \boldsymbol{J}_\xi^T\right)^{-1} \boldsymbol{J}_\xi,\ \ \ \ \boldsymbol{J}_\xi \equiv \nabla_r \boldsymbol{\xi}_r = \begin{pmatrix}
\partial_{r_1} \boldsymbol{\xi}_r \cdots \partial_{r_{3n}} \boldsymbol{\xi}_r
\end{pmatrix}.
\end{equation}
Then integration by parts yields
\begin{equation}
\nabla_R \mathcal{F}
= \frac{1}{Z}\int d\vecr\,\delta(\boldsymbol{\xi}_r(\vecr) - \vecR) e^{-\beta \mathcal{H}(\boldsymbol{r})}\left(\boldsymbol{\Xi}_F \nabla_r \mathcal{H} - \frac{1}{\beta} \nabla_r \cdot \boldsymbol{\Xi}_F\right),
\end{equation}
where $\nabla_r \cdot \boldsymbol{\Xi}_F$ is the divergence taken over the $3n$ column indices and so is a $3N \times 1$ vector.

Unlike in the linear case \eqref{force2Lin}, there are now two terms in $\nabla_R \mathcal{F}$ because the Jacobian and, thus, $\boldsymbol{\Xi}_F$ are no longer assumed to be constant. This implies that the CG force is no longer obtained by applying a projection to the AA force. In the language of ensemble averages and the AA force $\mathbf{F}_{\text{AA}}$, the first derivative and corresponding CG force $\mathbf{F}_{\text{CG}}$ are
\begin{equation}
\nabla_R \mathcal{F} = -\expval{\boldsymbol{\Xi}_F \mathbf{F}_{\text{AA}} + \frac{1}{\beta}\nabla_r \cdot \boldsymbol{\Xi}_F}_R \implies \mathbf{F}_{\text{CG}} = \expval{\boldsymbol{\Xi}_F \mathbf{F}_{\text{AA}} + \frac{1}{\beta}\nabla_r \cdot \boldsymbol{\Xi}_F}_R.\label{ensembleForceGen}
\end{equation}

\subsection{Differentiating the coarse-grained force}

To derive the formula for the CG Hessian, it is helpful to compute a slightly more general derivative---that of the ensemble average of an arbitrary function $\mathcal{O}(\mathbf{r})$ of the atomic position $\mathbf{r}$. This is because $\nabla_R \mathcal{F}$ itself is an ensemble average. By definition, we have that
\begin{equation}
\expval{\mathcal{O}}_R = \frac{1}{Z} \int d\vecr\,\delta(\boldsymbol{\xi}_r(\vecr) - \vecR)e^{-\beta\mathcal{H}(\vecr)}\mathcal{O}(\mathbf{r}).
\end{equation}
By using the product rule, we write
\begin{equation}
\begin{split}
\nabla_R \expval{\mathcal{O}}_R &= -\frac{\nabla_R Z}{Z^2} \int d\vecr\,\delta(\boldsymbol{\xi}_r(\vecr) - \vecR)e^{-\beta\mathcal{H}(\vecr)}\mathcal{O}(\mathbf{r})\\&\qquad\quad+ \frac{1}{Z}\int d\vecr\,\nabla_R \delta(\boldsymbol{\xi}_r(\vecr) - \vecR)e^{-\beta\mathcal{H}(\vecr)}\mathcal{O}(\mathbf{r}).
\end{split}
\end{equation}
The first term can be rewritten in terms of the free energy and $\expval{\mathcal{O}}_R$. The second term can be further manipulated by invoking \eqref{deltaDerivsGen}. This leads to
\begin{equation}
\nabla_R \expval{\mathcal{O}}_R = \beta\,\nabla_R \mathcal{F} \expval{\mathcal{O}}_R - \frac{1}{Z}\int d\vecr\,\boldsymbol{\Xi}_F\nabla_r \delta(\boldsymbol{\xi}_r(\vecr) - \vecR)e^{-\beta\mathcal{H}(\vecr)}\mathcal{O}(\mathbf{r}).
\end{equation}
Using \eqref{ensembleForceGen} and the definition $\mathbf{F}_{\text{AA}} \equiv -\nabla_r \mathcal{H}$, we arrive at the following equation:
\begin{align}
\nabla_R \expval{\mathcal{O}}_R
&= \expval{\boldsymbol{\Xi}_F \nabla_r \mathcal{O}}_R + \beta\big(\expval{\boldsymbol{\Xi}_F \mathbf{F}_{\text{AA}} \mathcal{O}}_R - \expval{\boldsymbol{\Xi}_F \mathbf{F}_{\text{AA}}}_R \expval{\mathcal{O}}_R\big)\nonumber\\
&\qquad+ \expval{\left(\nabla_r \cdot \boldsymbol{\Xi}_F\right)\mathcal{O}}_R - \expval{\nabla_r \cdot \boldsymbol{\Xi}_F}_R \expval{\mathcal{O}}_R.
\end{align}
We can also formulate a vectorial version of this identity; indeed, this is how we obtain the CG Hessian. Specifically, if we consider $\boldsymbol{\mathcal{O}}$ to be a $D$-dimensional vector represented as a $D \times 1$ matrix, then we write
\begin{align}
\nabla_R \expval{\boldsymbol{\mathcal{O}}}_R
&= \langle{\boldsymbol{\Xi}_F \nabla_r \boldsymbol{\mathcal{O}}^T}\rangle_R + \beta\big(\langle{\boldsymbol{\Xi}_F \mathbf{F}_{\text{AA}} \boldsymbol{\mathcal{O}}^T}\rangle_R - \expval{\boldsymbol{\Xi}_F \mathbf{F}_{\text{AA}}}_R \langle\boldsymbol{\mathcal{O}}^T\rangle_R\big)\nonumber\\
&\qquad+ \langle{\left(\nabla_r \cdot \boldsymbol{\Xi}_F\right)\boldsymbol{\mathcal{O}}}^T\rangle_R - \expval{\nabla_r \cdot \boldsymbol{\Xi}_F}_R \langle{\boldsymbol{\mathcal{O}}^T}\rangle_R,
\end{align}
with the result treated as a $3N \times D$ object. To compute the Hessian, we make the following substitution (taking $D = 3N$):
\begin{equation}
\boldsymbol{\mathcal{O}} \to -\left(\boldsymbol{\Xi}_F \mathbf{F}_{\text{AA}} + \frac{1}{\beta}\nabla_r \cdot \boldsymbol{\Xi}_F\right).
\end{equation}
After a tedious calculation using the fact that the AA Hessian $\mathbf{H}_{\text{AA}} \equiv -\nabla_r \mathbf{F}_{\text{AA}}$ is symmetric ($\mathbf{H}_{\text{AA}}^T = \mathbf{H}_{\text{AA}}$), we get the following generalization of the linear coarse-graining formula \eqref{eq:cg_hessian_main}:
\begin{align}
\mathbf{H}_{\text{CG}}
&= \langle{\boldsymbol{\Xi}_F \mathbf{H}_{\text{AA}}\boldsymbol{\Xi}_F^T}\rangle_R - \beta \boldsymbol{\Sigma}(\boldsymbol{\Xi}_F \mathbf{F}_{\text{AA}},\boldsymbol{\Xi}_F\mathbf{F}_{\text{AA}})\nonumber\\
&\qquad - \expval{\boldsymbol{\mathcal{T}}}_R - \boldsymbol{\Sigma}(\boldsymbol{\Xi}_F \mathbf{F}_{\text{AA}},\nabla_r \cdot \boldsymbol{\Xi}_F) - \boldsymbol{\Sigma}(\nabla_r \cdot \boldsymbol{\Xi}_F,\boldsymbol{\Xi}_F \mathbf{F}_{\text{AA}})\nonumber\\
&\qquad -\frac{1}{\beta} \left[\langle{\boldsymbol{\Xi}_F \nabla_r \left(\nabla_r \cdot \boldsymbol{\Xi}_F\right)^T}\rangle_R + \boldsymbol{\Sigma}(\nabla_r \cdot \boldsymbol{\Xi}_F,\nabla_r \cdot \boldsymbol{\Xi}_F)\right].\label{hessianGen}
\end{align}
where $\boldsymbol{\Sigma}(\cdot,\cdot)$ is the covariance matrix, and we define $\boldsymbol{\mathcal{T}}$ as a matrix with components
\begin{equation}
\left(\boldsymbol{\mathcal{T}}\right)_{IJ} = \sum_{i,j} \left(\mathbf{F}_{\text{AA}}\right)^{i}\left(\boldsymbol{\Xi}_F\right)_{Ij}\pdv{\left(\boldsymbol{\Xi}_F\right)_{Ji}}{r_j}.
\end{equation}
The first line of \eqref{hessianGen} is the same as in the linear case and reduces to it. The remaining terms all stem from the nonlinearity of $\boldsymbol{\xi}_r$ and vanish if $\boldsymbol{\Xi}_F$ is constant.

As a sanity check, we prove that the right-hand side of \eqref{hessianGen} is symmetric. The entire first line, the sum of covariances on the second line, and the covariance on the third line are all manifestly symmetric, so the only problematic piece is $-\expval{\mathcal{T}}_R - \beta^{-1} \langle{\boldsymbol{\Xi}_F \nabla_r \left(\nabla_r \cdot \boldsymbol{\Xi}_F\right)^T}\rangle_R$.

To manipulate this sum, we rewrite it in integral form and integrate by parts. Doing so in index notation (used for clarity) yields
\begin{align}\label{asymmetricQ}
&\left(-\expval{\boldsymbol{\mathcal{T}}}_R - \frac{1}{\beta}\langle{\boldsymbol{\Xi}_F \nabla_r \left(\nabla_r \cdot \boldsymbol{\Xi}_F\right)^T}\rangle_R\right)_{IJ}\\
&\ \ = \frac{1}{\beta Z} \sum_{i,j} \int d\vecr\,\left[\frac{\partial}{\partial r_i}\delta(\boldsymbol{\xi}_r(\vecr) - \vecR)(\boldsymbol{\Xi}_F)_{Ij}\pdv{(\boldsymbol{\Xi}_F)_{Ji}}{r_j} + \delta(\boldsymbol{\xi}_r(\vecr) - \vecR) \pdv{(\boldsymbol{\Xi}_F)_{Ij}}{r_i}\pdv{(\boldsymbol{\Xi}_F)_{Ji}}{r_j}\right]e^{-\beta \mathcal{H}(\vecr)}.\nonumber
\end{align}
The second term in the brackets is symmetric with respect to the free indices $IJ$, since we can freely swap the summation indices $ij$. As for the first term, we must use the fact that
\begin{equation}
\frac{\partial}{\partial r_i}\delta(\boldsymbol{\xi}_r(\vecr) - \vecR) = -\sum_K (\mathbf{J}_\xi)_{Ki} \frac{\partial}{\partial R_K}\delta(\boldsymbol{\xi}_r(\vecr) - \vecR),\label{deltarInv}
\end{equation}
as well as the following identity (where $\delta_{JK}$ here is the Kronecker delta):
\begin{equation}
\sum_i (\boldsymbol{\Xi}_F)_{Ji}(\mathbf{J}_\xi)_{Ki} = \delta_{JK}.\label{XiJId}
\end{equation}
This is simply the equation $\boldsymbol{\Xi}_F \mathbf{J}_\xi^T = \mathbf{I}_{3N}$ in index form. \eqref{XiJId} can be differentiated to obtain
\begin{equation}
\sum_i \frac{\partial(\boldsymbol{\Xi}_F)_{Ji}}{\partial r_j} (\mathbf{J}_\xi)_{Ki} + \sum_i (\boldsymbol{\Xi}_F)_{Ji} \pdv{(\mathbf{J}_\xi)_{Ki}}{r_j} = 0.\label{diffJXi}
\end{equation}
By making use of \eqref{deltarInv} and \eqref{diffJXi} in \eqref{asymmetricQ}, the first term becomes
\begin{equation}
\frac{1}{\beta Z}\sum_{i,j,K} \frac{\partial}{\partial R_K} \int d\vecr\,\delta(\boldsymbol{\xi}_r(\vecr) - \vecR)e^{-\beta \mathcal{H}(\vecr)} \left(\frac{\partial^2}{\partial r_i\,\partial r_j}(\boldsymbol{\xi}_r)_K\right) (\boldsymbol{\Xi}_F)_{Ij} (\boldsymbol{\Xi}_F)_{Ji},
\end{equation}
where $(\boldsymbol{\xi}_r)_K$ is the $K$th component of the coarse-graining map. Just like the second term in \eqref{asymmetricQ}, this expression is symmetric with respect to the free indices, so the proof is complete.
\newpage

\section{Supplemental material for in-distribution models}\label{app:inDistSupplement}

In the main paper (Table~\ref{tab:id_combined}) we report the two slowest modes together with representative structural metrics for the three variants trained on the benchmark proteins of \citet{Aghili2025}. This appendix provides the remaining results for those models, including the full set of slow-mode metrics across all four leading TICA components (Table~\ref{tab:id_slow}), the complete set of structural metrics including bond-length, bond-angle, radius-of-gyration, and dihedral distributions (Table~\ref{tab:id_structural}), the training and validation loss curves (Figures~\ref{fig:train_loss_id}--\ref{fig:val_loss_id}), and the learned probability density functions for two proteins, chignolin (Figures~\ref{fig:chignolin2D_id}--\ref{fig:chignolinStructure_id}) and $\alpha$3D (Figures~\ref{fig:a3d2D_id}--\ref{fig:a3dStructure_id}).

\subsection{Slow-mode metrics}

Table~\ref{tab:id_slow} reports the full set of slow-mode metrics. In addition to the metrics shown in the main paper, which are reproduced here for convenience, we show KL divergences of distributions along TIC2 and TIC3. We also include the W1 distance for the TICA 2D distribution, which is computed via Gaussian Kernel density estimation along the two slowest components TIC0--1. The HVP-matched models win over force matching even on most of these additional metrics.


\begin{table}[h]
\centering
\setlength{\tabcolsep}{6pt}
\begin{tabular}{@{}l@{\hspace{6pt}}lrrrrr@{}}
\toprule
\multicolumn{7}{c}{\textbf{Slow-mode metrics}} \\
\midrule
Protein & Model & \makecell{TICA 2D\\W1 $\downarrow$} & \makecell{TIC 0\\KL $\downarrow$} & \makecell{TIC 1\\KL $\downarrow$} & \makecell{TIC 2\\KL $\downarrow$} & \makecell{TIC 3\\KL $\downarrow$} \\
\midrule
\multirow{3}{*}{\shortstack[l]{Chignolin\\ $N=10$}} & FM & $1.016_{\pm 0.052}$ & $2.23_{\pm 0.53}$ & $1.64_{\pm 0.53}$ & $0.22_{\pm 0.07}$ & $0.12_{\pm 0.05}$ \\
 & +AA$_{\text{p}}$ & $\mathbf{0.640_{\pm 0.035}}$ & $\mathbf{0.40_{\pm 0.11}}$ & $\mathbf{0.64_{\pm 0.13}}$ & $0.14_{\pm 0.02}$ & $0.12_{\pm 0.05}$ \\
 & +Cov & $0.728_{\pm 0.019}$ & $1.23_{\pm 0.24}$ & $0.78_{\pm 0.05}$ & $\mathbf{0.11_{\pm 0.03}}$ & $\mathbf{0.09_{\pm 0.03}}$ \\
\midrule
\multirow{3}{*}{\shortstack[l]{Trp-cage\\ $N=20$}} & FM & $0.572_{\pm 0.086}$ & $0.94_{\pm 0.39}$ & $1.04_{\pm 0.45}$ & $0.65_{\pm 0.34}$ & $0.47_{\pm 0.27}$ \\
 & +AA$_{\text{p}}$ & $0.631_{\pm 0.057}$ & $1.05_{\pm 0.05}$ & $1.18_{\pm 0.28}$ & $0.82_{\pm 0.24}$ & $\mathbf{0.32_{\pm 0.14}}$ \\
 & +Cov & $\mathbf{0.506_{\pm 0.045}}$ & $\mathbf{0.78_{\pm 0.10}}$ & $\mathbf{0.98_{\pm 0.06}}$ & $\mathbf{0.37_{\pm 0.11}}$ & $0.52_{\pm 0.11}$ \\
\midrule
\multirow{3}{*}{\shortstack[l]{BBA\\ $N=28$}} & FM & $1.088_{\pm 0.068}$ & $\mathbf{2.90_{\pm 1.08}}$ & $\mathbf{0.60_{\pm 0.37}}$ & $\mathbf{0.76_{\pm 0.34}}$ & $0.53_{\pm 0.07}$ \\
 & +AA$_{\text{p}}$ & $\mathbf{0.971_{\pm 0.056}}$ & $4.78_{\pm 0.76}$ & $0.72_{\pm 0.40}$ & $0.95_{\pm 0.41}$ & $\mathbf{0.46_{\pm 0.12}}$ \\
 & +Cov & $0.971_{\pm 0.051}$ & $3.91_{\pm 0.32}$ & $0.82_{\pm 0.26}$ & $0.76_{\pm 0.17}$ & $0.68_{\pm 0.27}$ \\
\midrule
\multirow{3}{*}{\shortstack[l]{WW domain\\ $N=34$}} & FM & $\mathbf{0.949_{\pm 0.176}}$ & $2.05_{\pm 1.06}$ & $1.41_{\pm 0.14}$ & $2.31_{\pm 0.49}$ & $\mathbf{1.42_{\pm 0.18}}$ \\
 & +AA$_{\text{p}}$ & $1.082_{\pm 0.102}$ & $2.92_{\pm 1.87}$ & $1.65_{\pm 0.36}$ & $\mathbf{2.07_{\pm 0.48}}$ & $1.83_{\pm 0.63}$ \\
 & +Cov & $0.985_{\pm 0.137}$ & $\mathbf{1.66_{\pm 0.43}}$ & $\mathbf{1.30_{\pm 0.19}}$ & $2.35_{\pm 0.57}$ & $2.09_{\pm 0.48}$ \\
\midrule
\multirow{3}{*}{\shortstack[l]{Protein B\\ $N=47$}} & FM & $1.120_{\pm 0.159}$ & $3.50_{\pm 2.19}$ & $4.74_{\pm 3.21}$ & $1.95_{\pm 1.89}$ & $3.07_{\pm 1.79}$ \\
 & +AA$_{\text{p}}$ & $1.009_{\pm 0.435}$ & $\mathbf{1.77_{\pm 1.14}}$ & $\mathbf{0.60_{\pm 0.34}}$ & $\mathbf{0.51_{\pm 0.18}}$ & $\mathbf{0.60_{\pm 0.38}}$ \\
 & +Cov & $\mathbf{0.615_{\pm 0.029}}$ & $1.96_{\pm 0.39}$ & $0.81_{\pm 0.41}$ & $0.87_{\pm 0.30}$ & $1.62_{\pm 0.86}$ \\
\midrule
\multirow{3}{*}{\shortstack[l]{Homeodomain\\ $N=54$}} & FM & $\mathbf{0.864_{\pm 0.104}}$ & $4.50_{\pm 1.97}$ & $3.93_{\pm 0.81}$ & $\mathbf{2.44_{\pm 1.20}}$ & $1.33_{\pm 0.57}$ \\
 & +AA$_{\text{p}}$ & $0.896_{\pm 0.099}$ & $\mathbf{2.33_{\pm 0.50}}$ & $\mathbf{2.24_{\pm 0.90}}$ & $2.65_{\pm 0.61}$ & $\mathbf{0.69_{\pm 0.08}}$ \\
 & +Cov & $0.954_{\pm 0.054}$ & $2.50_{\pm 1.16}$ & $3.29_{\pm 1.97}$ & $3.34_{\pm 0.94}$ & $0.85_{\pm 0.12}$ \\
\midrule
\multirow{3}{*}{\shortstack[l]{Protein G\\ $N=56$}} & FM & $1.072_{\pm 0.144}$ & $3.62_{\pm 0.65}$ & $1.93_{\pm 0.30}$ & $\mathbf{0.89_{\pm 0.03}}$ & $1.53_{\pm 0.47}$ \\
 & +AA$_{\text{p}}$ & $1.132_{\pm 0.139}$ & $3.35_{\pm 0.81}$ & $2.20_{\pm 0.29}$ & $0.92_{\pm 0.16}$ & $2.08_{\pm 0.71}$ \\
 & +Cov & $\mathbf{1.049_{\pm 0.228}}$ & $\mathbf{2.74_{\pm 1.53}}$ & $\mathbf{1.56_{\pm 0.71}}$ & $0.91_{\pm 0.09}$ & $\mathbf{1.43_{\pm 0.55}}$ \\
\midrule
\multirow{3}{*}{\shortstack[l]{$\alpha$3D\\ $N=73$}} & FM & $0.749_{\pm 0.128}$ & $1.89_{\pm 1.99}$ & $2.77_{\pm 0.75}$ & $1.39_{\pm 0.54}$ & $1.93_{\pm 1.03}$ \\
 & +AA$_{\text{p}}$ & $0.707_{\pm 0.145}$ & $1.75_{\pm 1.29}$ & $1.71_{\pm 0.79}$ & $1.66_{\pm 0.52}$ & $1.65_{\pm 0.83}$ \\
 & +Cov & $\mathbf{0.623_{\pm 0.067}}$ & $\mathbf{0.67_{\pm 0.16}}$ & $\mathbf{0.90_{\pm 0.24}}$ & $\mathbf{0.88_{\pm 0.22}}$ & $\mathbf{1.01_{\pm 0.38}}$ \\
\midrule
\multirow{3}{*}{\shortstack[l]{$\lambda$-repressor\\ $N=80$}} & FM & $0.964_{\pm 0.187}$ & $5.83_{\pm 1.96}$ & $4.42_{\pm 3.36}$ & $3.37_{\pm 1.37}$ & $3.68_{\pm 1.42}$ \\
 & +AA$_{\text{p}}$ & $\mathbf{0.887_{\pm 0.270}}$ & $\mathbf{3.84_{\pm 2.42}}$ & $3.79_{\pm 2.03}$ & $3.35_{\pm 2.07}$ & $3.70_{\pm 2.44}$ \\
 & +Cov & $0.999_{\pm 0.074}$ & $5.88_{\pm 1.04}$ & $\mathbf{2.85_{\pm 0.92}}$ & $\mathbf{2.18_{\pm 0.45}}$ & $\mathbf{2.60_{\pm 0.64}}$ \\
\bottomrule
\end{tabular}
\caption{In-distribution benchmark, consisting of W1 distance on estimated density over TIC0--1 and KL divergences on TIC0--3. Metrics are computed over 5 seeds (mean$_{\pm\text{std}}$), with best per protein in bold; lower mean is better.}
\label{tab:id_slow}
\end{table}

\subsection{Structural metrics}

Table~\ref{tab:id_structural} reports the full set of structural W1 distances. As stated in the main text, the clearest structural improvement is on bond lengths, in which an HVP-matching variant achieves the lowest bond-length W1 on eight of nine proteins, often by a large margin. For example, chignolin improves from $1.1\times10^{-3}$ under FM to $3\times10^{-4}$, and $\alpha$3D from $7.3\times10^{-3}$ to $6\times10^{-4}$, which is consistent with the Hessian directly governing bonded stiffness. Bond-angle and dihedral improvements are smaller and protein-dependent, without a consistent ordering across variants, as expected for local geometric properties that force matching already constrains reasonably well.

\begin{table}[h]

\centering
\setlength{\tabcolsep}{6pt}

\begin{tabular}{@{}l@{\hspace{6pt}}lrrrr@{}}
\toprule
\multicolumn{6}{c}{\textbf{Structural metrics (W1)}} \\
\midrule
Protein & Model & \makecell{Length\\$(\times 10^{-2})$ $\downarrow$} & Angle $\downarrow$ & RoG $\downarrow$ & Dihedral $\downarrow$ \\
\midrule
\multirow{3}{*}{\shortstack[l]{Chignolin\\ $N=10$}}
 & FM            & $0.11_{\pm 0.03}$ & $0.080_{\pm 0.010}$ & $0.171_{\pm 0.005}$ & $0.618_{\pm 0.042}$ \\
 & +AA$_{\text{p}}$     & $\mathbf{0.03_{\pm 0.01}}$ & $0.060_{\pm 0.009}$ & $0.132_{\pm 0.009}$ & $0.445_{\pm 0.088}$ \\
 & +Cov & $\mathbf{0.03_{\pm 0.01}}$ & $\mathbf{0.033_{\pm 0.005}}$ & $\mathbf{0.120_{\pm 0.002}}$ & $\mathbf{0.329_{\pm 0.048}}$ \\
\midrule
\multirow{3}{*}{\shortstack[l]{Trp-cage\\ $N=20$}}
 & FM            & $0.83_{\pm 0.17}$ & $0.096_{\pm 0.020}$ & $\mathbf{0.107_{\pm 0.025}}$ & $0.155_{\pm 0.022}$ \\
 & +AA$_{\text{p}}$     & $\mathbf{0.04_{\pm 0.02}}$ & $\mathbf{0.063_{\pm 0.011}}$ & $0.108_{\pm 0.007}$ & $0.205_{\pm 0.045}$ \\
 & +Cov & $0.04_{\pm 0.01}$ & $0.065_{\pm 0.008}$ & $0.134_{\pm 0.009}$ & $\mathbf{0.147_{\pm 0.012}}$ \\
\midrule
\multirow{3}{*}{\shortstack[l]{BBA\\ $N=28$}}
 & FM            & $1.96_{\pm 1.55}$ & $0.056_{\pm 0.010}$ & $0.325_{\pm 0.237}$ & $\mathbf{0.243_{\pm 0.083}}$ \\
 & +AA$_{\text{p}}$     & $0.29_{\pm 0.16}$ & $\mathbf{0.051_{\pm 0.011}}$ & $0.254_{\pm 0.031}$ & $0.258_{\pm 0.043}$ \\
 & +Cov & $\mathbf{0.07_{\pm 0.02}}$ & $0.081_{\pm 0.014}$ & $\mathbf{0.236_{\pm 0.034}}$ & $0.287_{\pm 0.035}$ \\
\midrule
\multirow{3}{*}{\shortstack[l]{WW domain\\ $N=34$}}
 & FM            & $\mathbf{0.13_{\pm 0.06}}$ & $\mathbf{0.075_{\pm 0.008}}$ & $0.199_{\pm 0.066}$ & $0.519_{\pm 0.031}$ \\
 & +AA$_{\text{p}}$     & $0.64_{\pm 0.09}$ & $0.084_{\pm 0.004}$ & $0.259_{\pm 0.018}$ & $0.538_{\pm 0.035}$ \\
 & +Cov & $0.31_{\pm 0.14}$ & $0.096_{\pm 0.010}$ & $\mathbf{0.195_{\pm 0.011}}$ & $\mathbf{0.497_{\pm 0.048}}$ \\
\midrule
\multirow{3}{*}{\shortstack[l]{Protein B\\ $N=47$}}
 & FM            & $0.19_{\pm 0.31}$ & $0.153_{\pm 0.051}$ & $0.167_{\pm 0.035}$ & $0.300_{\pm 0.106}$ \\
 & +AA$_{\text{p}}$     & $0.94_{\pm 1.08}$ & $\mathbf{0.072_{\pm 0.015}}$ & $0.197_{\pm 0.014}$ & $0.248_{\pm 0.077}$ \\
 & +Cov & $\mathbf{0.04_{\pm 0.01}}$ & $0.134_{\pm 0.019}$ & $\mathbf{0.122_{\pm 0.015}}$ & $\mathbf{0.201_{\pm 0.012}}$ \\
\midrule
\multirow{3}{*}{\shortstack[l]{Homeodomain\\ $N=54$}}
 & FM            & $0.04_{\pm 0.01}$ & $0.045_{\pm 0.008}$ & $\mathbf{0.069_{\pm 0.025}}$ & $0.204_{\pm 0.037}$ \\
 & +AA$_{\text{p}}$     & $0.06_{\pm 0.02}$ & $0.044_{\pm 0.004}$ & $0.106_{\pm 0.019}$ & $\mathbf{0.167_{\pm 0.023}}$ \\
 & +Cov & $\mathbf{0.03_{\pm 0.01}}$ & $\mathbf{0.040_{\pm 0.002}}$ & $0.098_{\pm 0.005}$ & $0.189_{\pm 0.019}$ \\
\midrule
\multirow{3}{*}{\shortstack[l]{Protein G\\ $N=56$}}
 & FM            & $0.20_{\pm 0.07}$ & $0.075_{\pm 0.016}$ & $\mathbf{0.245_{\pm 0.022}}$ & $0.389_{\pm 0.041}$ \\
 & +AA$_{\text{p}}$     & $0.60_{\pm 0.11}$ & $0.089_{\pm 0.015}$ & $0.270_{\pm 0.013}$ & $\mathbf{0.357_{\pm 0.029}}$ \\
 & +Cov & $\mathbf{0.17_{\pm 0.02}}$ & $\mathbf{0.062_{\pm 0.004}}$ & $0.254_{\pm 0.025}$ & $0.373_{\pm 0.028}$ \\
\midrule
\multirow{3}{*}{\shortstack[l]{$\alpha$3D\\ $N=73$}}
 & FM            & $0.73_{\pm 1.49}$ & $0.043_{\pm 0.003}$ & $0.141_{\pm 0.030}$ & $0.207_{\pm 0.079}$ \\
 & +AA$_{\text{p}}$     & $\mathbf{0.06_{\pm 0.01}}$ & $0.046_{\pm 0.005}$ & $\mathbf{0.122_{\pm 0.013}}$ & $\mathbf{0.177_{\pm 0.023}}$ \\
 & +Cov & $0.09_{\pm 0.02}$ & $\mathbf{0.040_{\pm 0.004}}$ & $0.161_{\pm 0.024}$ & $0.357_{\pm 0.073}$ \\
\midrule
\multirow{3}{*}{\shortstack[l]{$\lambda$-repressor\\ $N=80$}}
 & FM            & $0.17_{\pm 0.05}$ & $0.066_{\pm 0.034}$ & $0.289_{\pm 0.084}$ & $0.235_{\pm 0.041}$ \\
 & +AA$_{\text{p}}$     & $0.18_{\pm 0.09}$ & $0.064_{\pm 0.013}$ & $0.240_{\pm 0.106}$ & $0.262_{\pm 0.081}$ \\
 & +Cov & $\mathbf{0.11_{\pm 0.01}}$ & $\mathbf{0.044_{\pm 0.002}}$ & $\mathbf{0.217_{\pm 0.030}}$ & $\mathbf{0.222_{\pm 0.026}}$ \\
\bottomrule
\end{tabular}
\caption{In-distribution benchmark, consisting of W1 distances on bond-length, bond-angle, radius-of-gyration, and dihedral distributions. Metrics are computed over 5 seeds (mean$_{\pm\text{std}}$), with best per protein in bold; lower mean is better.}
\label{tab:id_structural}
\end{table}
\newpage

\subsection{Training and validation loss curves}

\begin{figure}[H]
 \centering
 \includegraphics[scale=0.8]{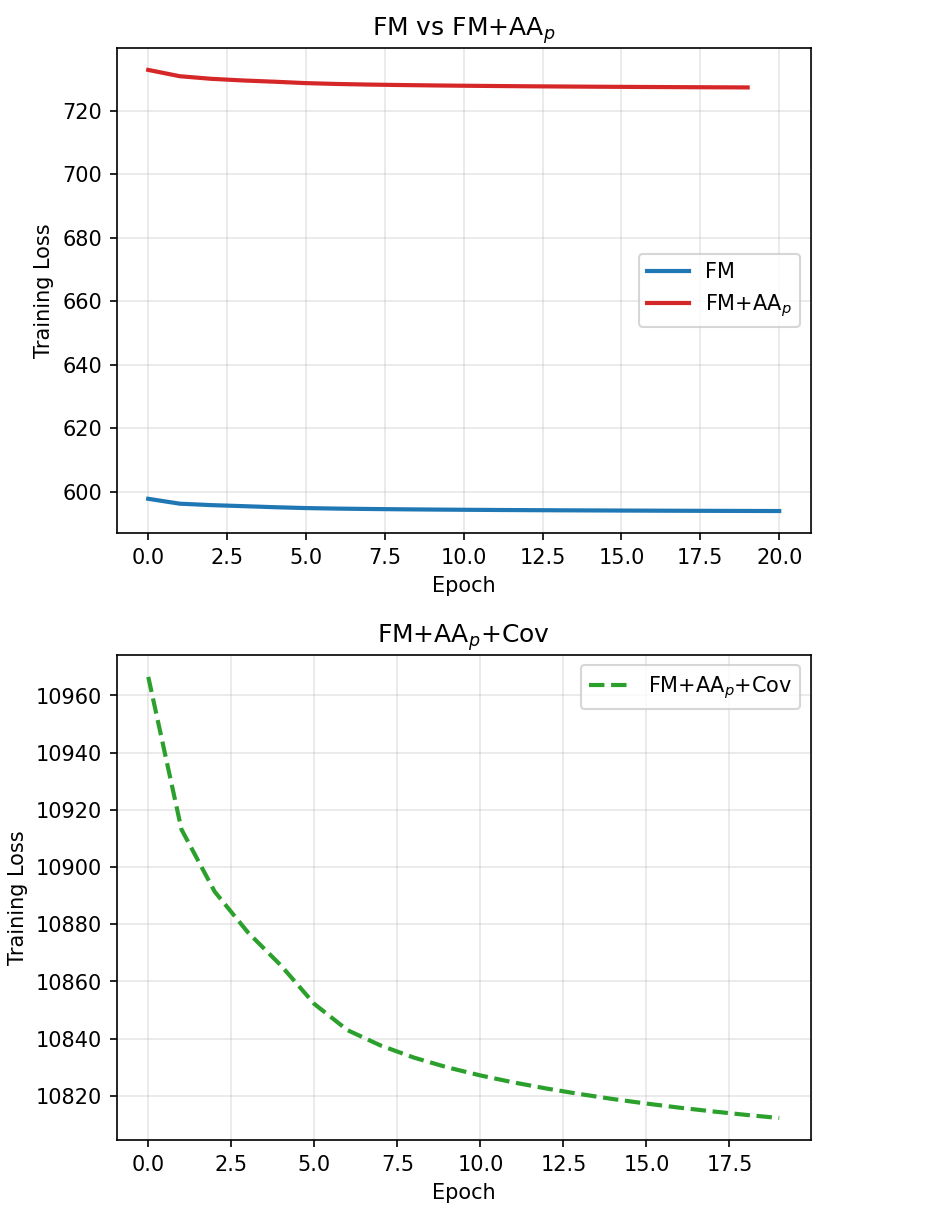}
  \caption{Training loss decomposition for the in-distribution models. \textbf{Top:}~FM and FM+AA$_\text{p}$ training losses, where the ${\sim}133$~kcal$^2$\,mol$^{-2}$\,\AA$^{-2}$ offset reflects the HVP matching term ($w_{\mathrm{HVP}} \|\mathbf{H}_{\mathrm{CG}}\mathbf{v} - \mathbf{H}_\theta\mathbf{v}\|^2$). \textbf{Bottom:}~FM+AA$_\text{p}$+Cov training loss, dominated by the covariance correction (Term~2). Despite the order-of-magnitude difference in training-loss scales, all three models achieve equivalent force-prediction accuracy on held-out data (Figure~\ref{fig:val_loss_id}).}
 \label{fig:train_loss_id}
 \end{figure}

\begin{figure}[H]
 \centering
 \includegraphics[scale=0.7]{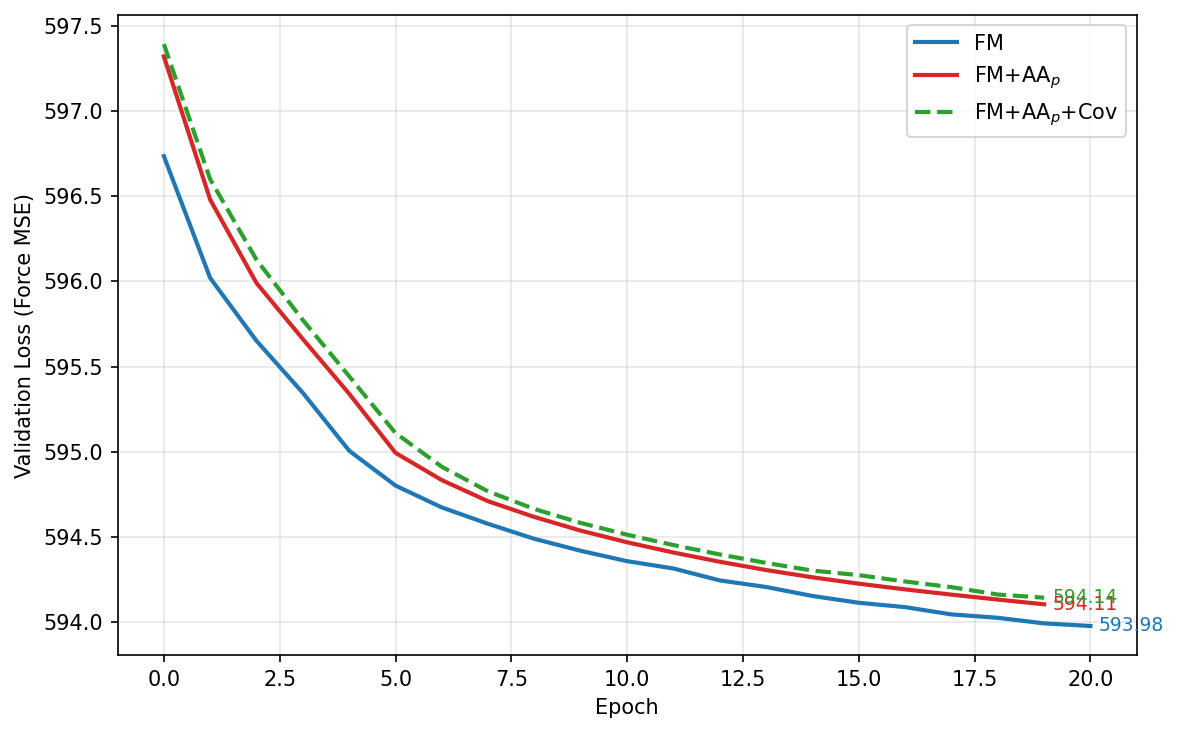}
 \caption{Validation loss (force matching MSE) during training on the nine benchmark proteins at $T=300$\,K. All three objectives (FM, FM+AA$_\text{p}$, and FM+AA$_\text{p}$+Cov) converge to comparable validation performance (${\sim}594$~kcal$^2$\,mol$^{-2}$\,\AA$^{-2}$). Models were trained for 21 (FM), 20 (FM+AA$_\text{p}$), and 20 (FM+AA$_\text{p}$+Cov) epochs with $w_\text{HVP}=0.01$ (for HVP-matching models) and batch size 200 on 4$\times$A40 GPUs.}
\label{fig:val_loss_id}
\end{figure}

We report loss curves for the in-distribution models. In Figure \ref{fig:train_loss_id}, we show the training losses versus epoch to visualize convergence. In Figure \ref{fig:val_loss_id}, we show the force matching mean squared errors (MSE) versus epoch to demonstrate that HVP-matching terms do not degrade force-prediction accuracy.

\newpage
 
\newpage

\subsection{Distributions}

\begin{figure}[h]
\centering
\subfloat[\textbf{FM}]{\includegraphics[scale=0.45]{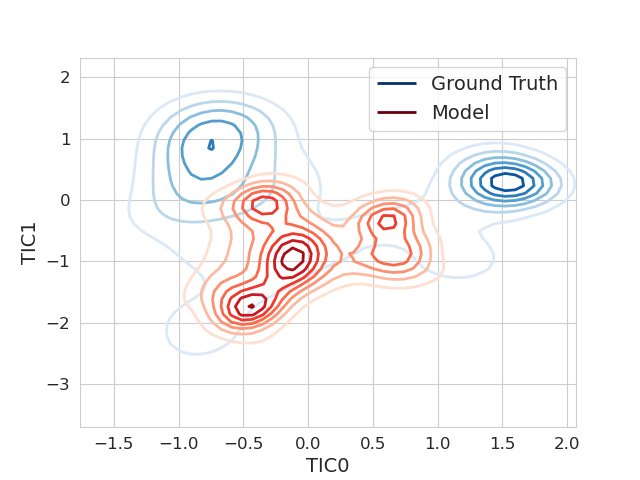}}\\
\subfloat[\textbf{FM+AA$_{\text{p}}$}]{\includegraphics[scale=0.45]{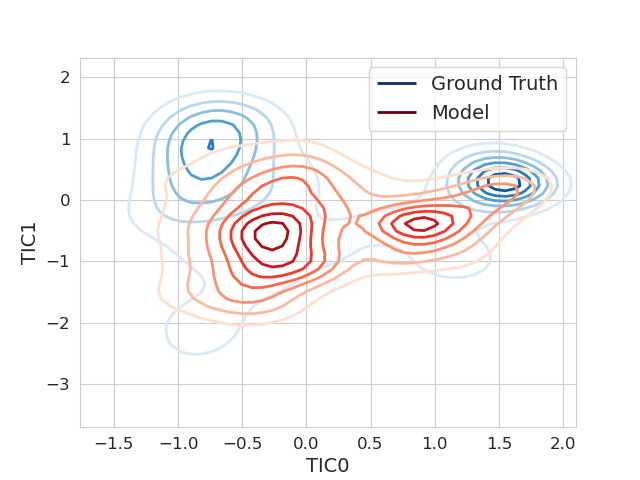}}
\subfloat[\textbf{FM+AA$_{\text{p}}$+Cov}]{\includegraphics[scale=0.45]{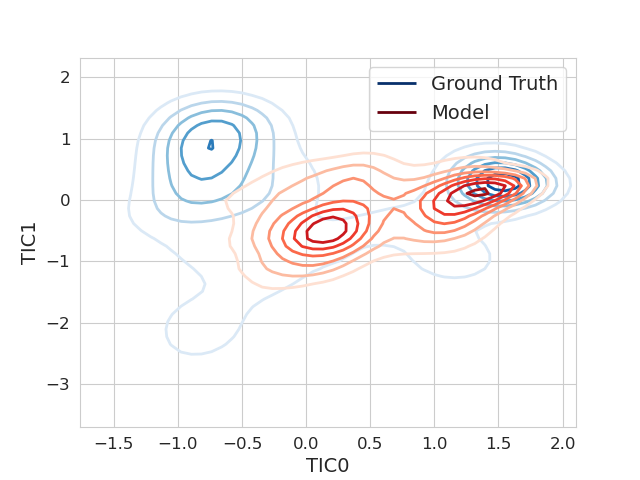}}
\caption{TICA 2D distributions for chignolin showing the Gaussian kernel density estimates of the free energy projected along TIC0 and TIC1. W1 distances between each model distribution and ground-truth distribution are reported in Table \ref{tab:id_slow}.}
\label{fig:chignolin2D_id}
\end{figure}

\begin{figure}
\centering
\subfloat[\textbf{FM}]{\includegraphics[scale=0.425]{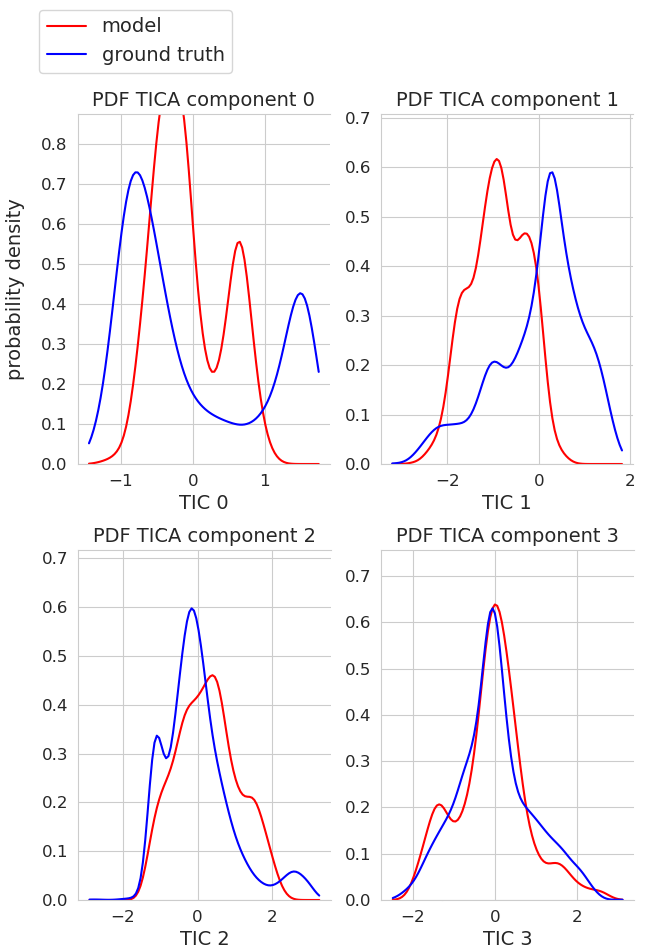}}\\
\subfloat[\textbf{FM+AA$_{\text{p}}$}]{\includegraphics[scale=0.425]{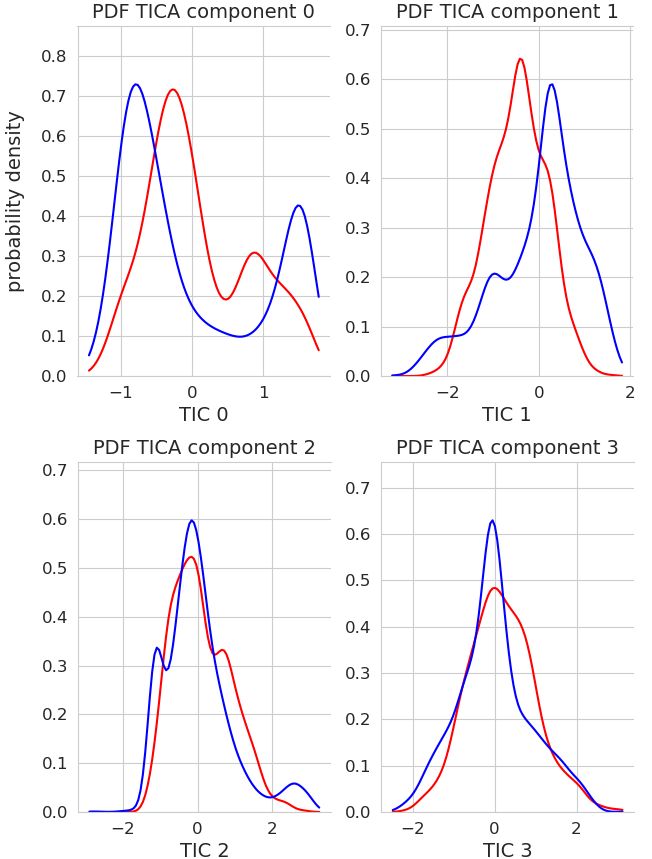}}
\subfloat[\textbf{FM+AA$_{\text{p}}$+Cov}]{\includegraphics[scale=0.425]{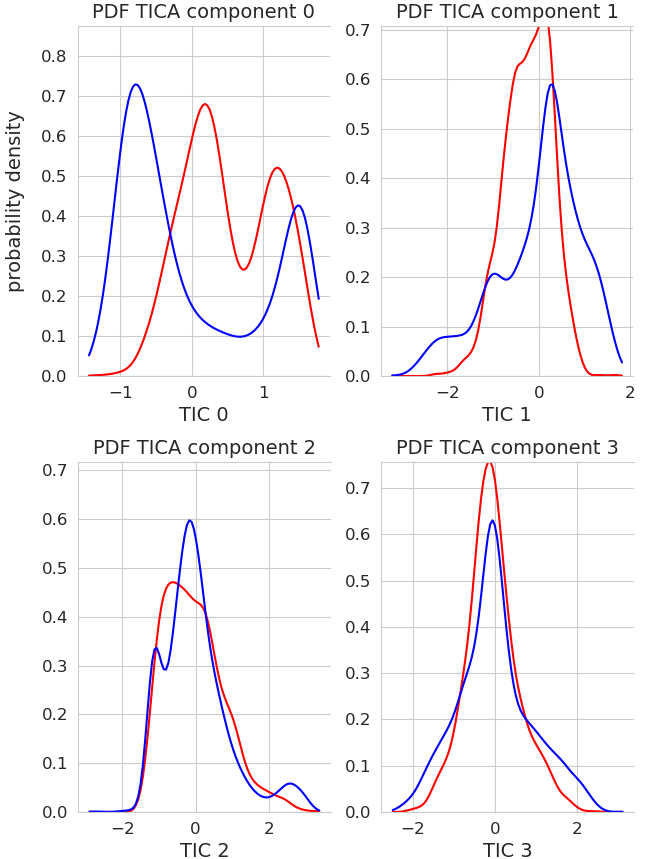}}
\caption{Slow-mode distributions for chignolin showing TIC0--3 projections of the free energy. The KL divergences between each model distribution and ground-truth distribution are reported in Table \ref{tab:id_slow}.}
\label{fig:chignolinTICA_id}
\end{figure}

\begin{figure}
\centering
\subfloat[\textbf{FM}]{\includegraphics[scale=0.5]{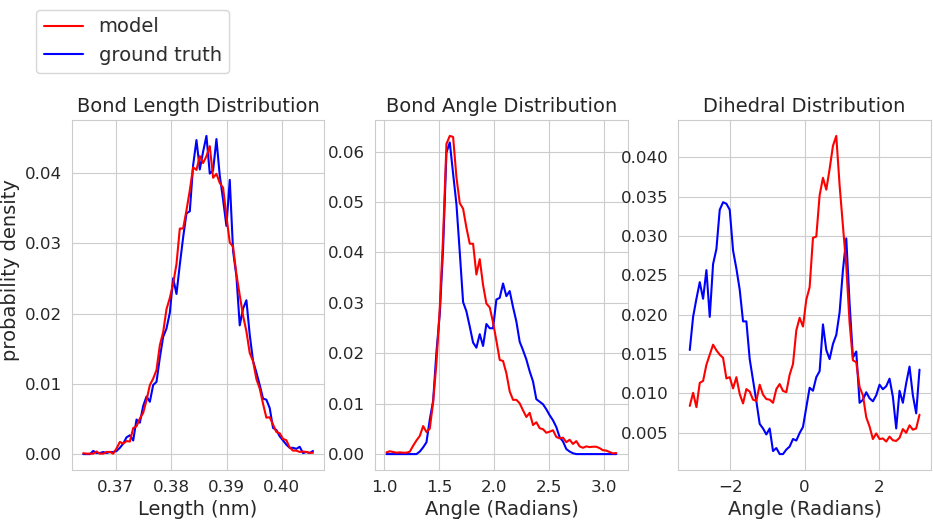}}\\
\subfloat[\textbf{FM+AA$_{\text{p}}$}]{\includegraphics[scale=0.5]{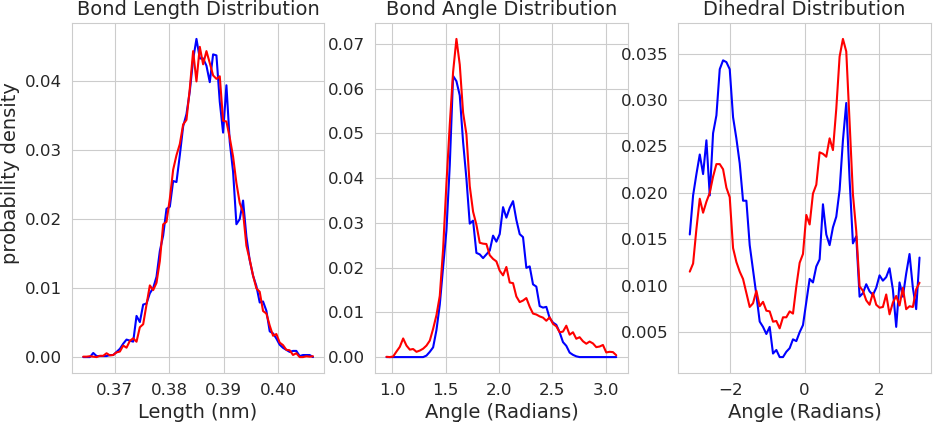}}\\
\subfloat[\textbf{FM+AA$_{\text{p}}$+Cov}]{\includegraphics[scale=0.5]{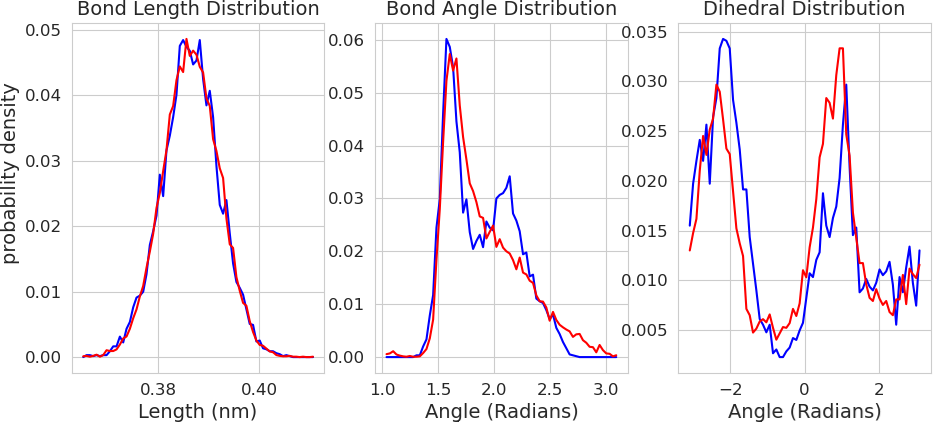}}
\caption{Structural distributions for chignolin showing bond lengths, bond angles, and dihedral angles. The W1 distances between each model distribution and ground-truth distribution are reported in Table \ref{tab:id_structural}.}
\label{fig:chignolinStructure_id}
\end{figure}

\begin{figure}
\centering
\subfloat[\textbf{FM}]{\includegraphics[scale=0.45]{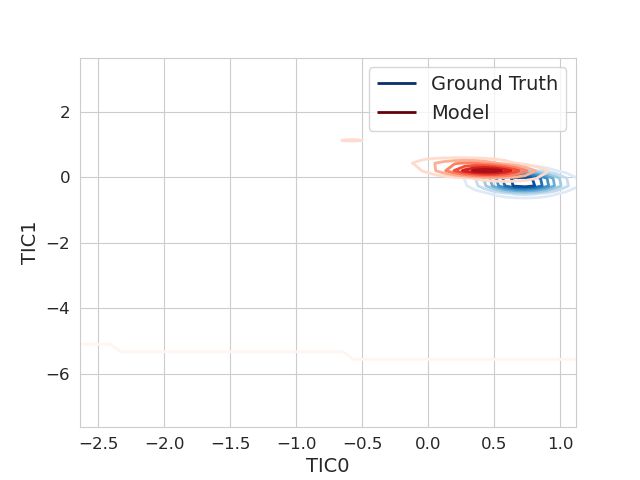}}\\
\subfloat[\textbf{FM+AA$_{\text{p}}$}]{\includegraphics[scale=0.45]{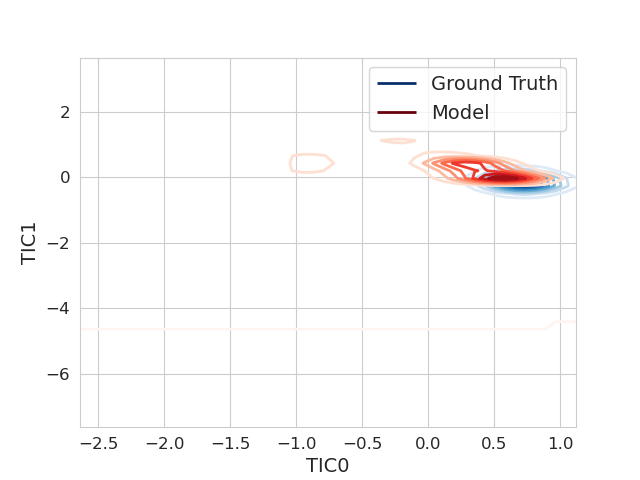}}
\subfloat[\textbf{FM+AA$_{\text{p}}$+Cov}]{\includegraphics[scale=0.45]{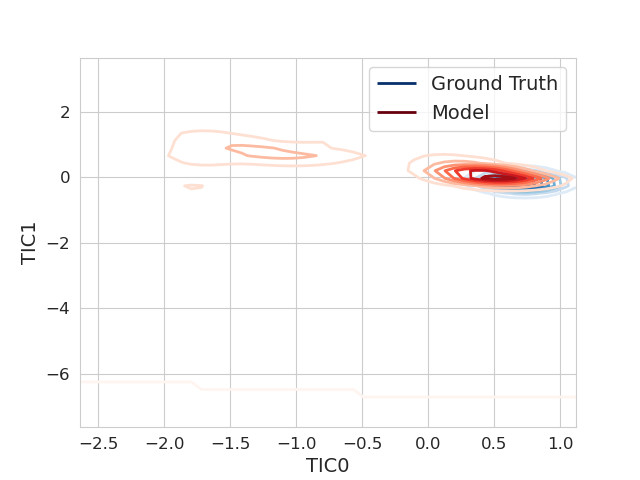}}
\caption{TICA 2D distributions for $\alpha$3D showing the Gaussian kernel density estimate of the free energy projected along TIC0 and TIC1. W1 distances between each model distribution and ground-truth distribution are reported in Table \ref{tab:id_slow}.}
\label{fig:a3d2D_id}
\end{figure}

\begin{figure}
\centering
\subfloat[\textbf{FM}]{\includegraphics[scale=0.425]{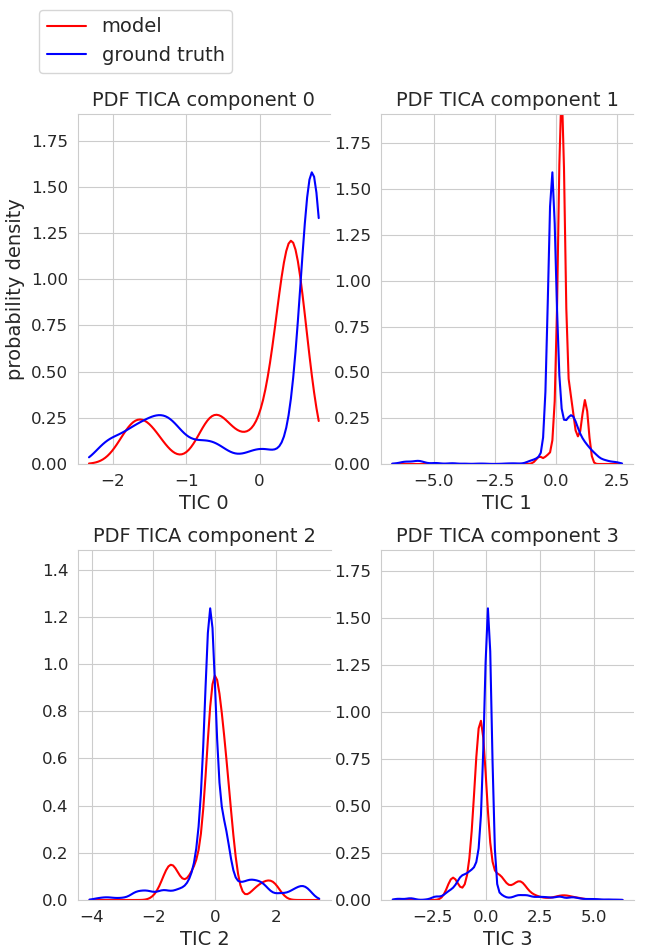}}\\
\subfloat[\textbf{FM+AA$_{\text{p}}$}]{\includegraphics[scale=0.425]{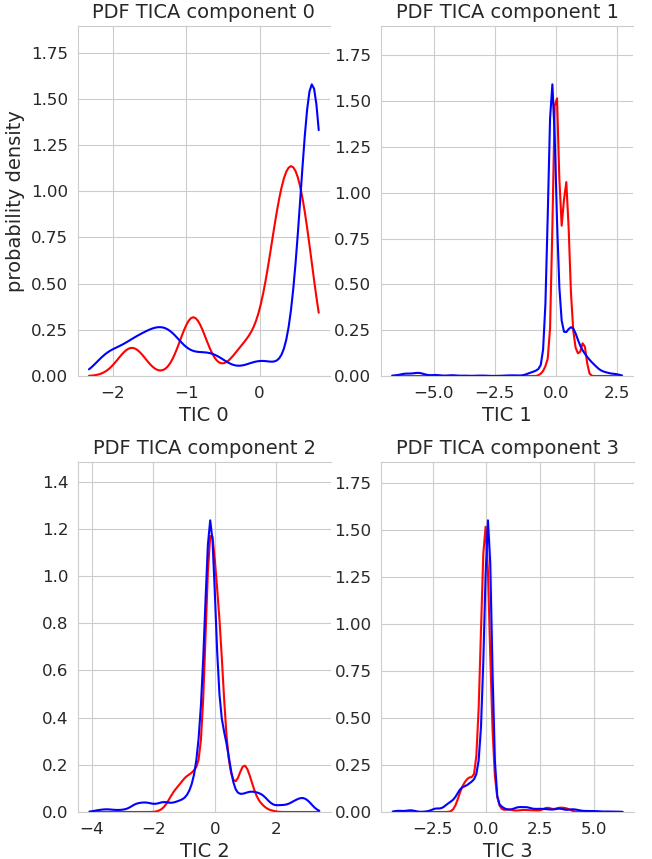}}
\subfloat[\textbf{FM+AA$_{\text{p}}$+Cov}]{\includegraphics[scale=0.425]{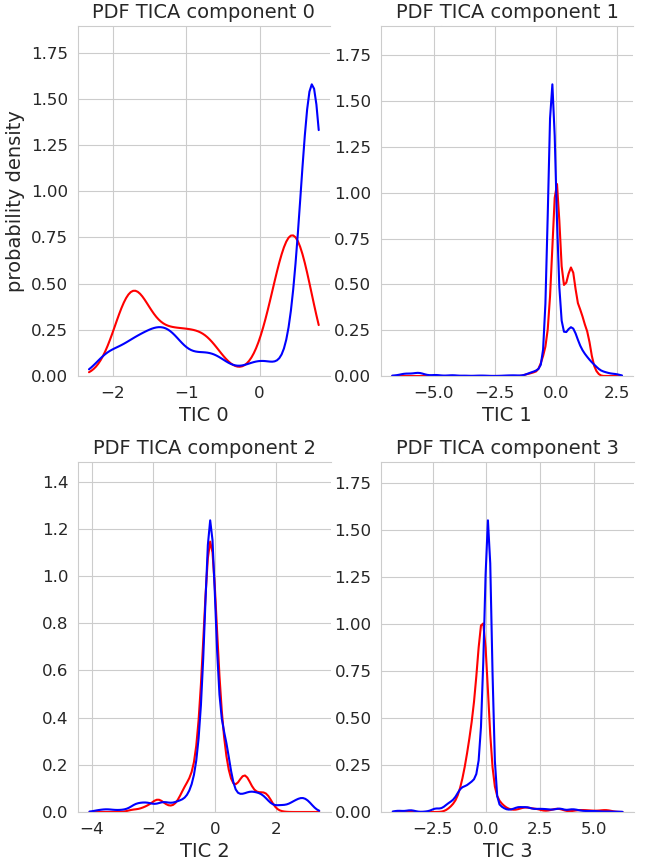}}
\caption{Slow-mode distributions for $\alpha$3D showing TIC0--3 projections of the free energy. The KL divergences between each model distribution and ground-truth distribution are reported in Table \ref{tab:id_slow}.}
\label{fig:a3dTICA_id}
\end{figure}

\begin{figure}
\centering
\subfloat[\textbf{FM}]{\includegraphics[scale=0.5]{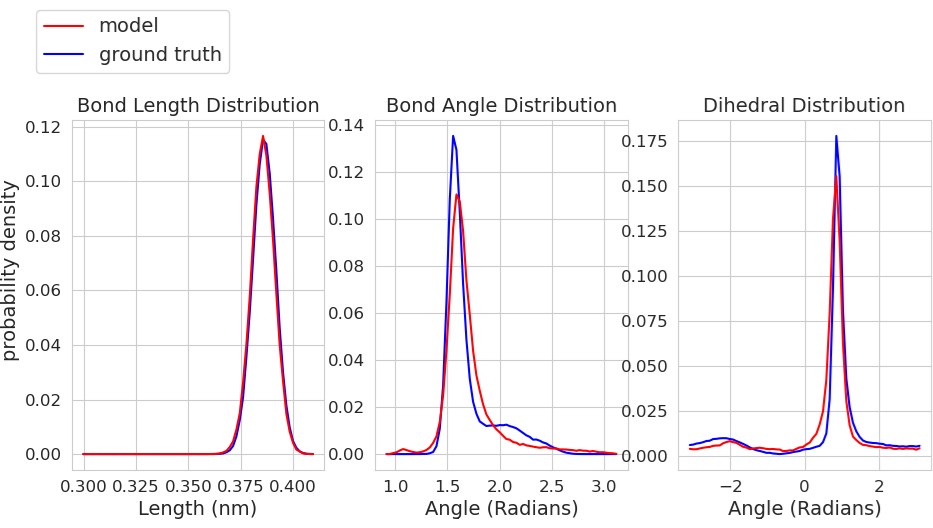}}\\
\subfloat[\textbf{FM+AA$_{\text{p}}$}]{\includegraphics[scale=0.5]{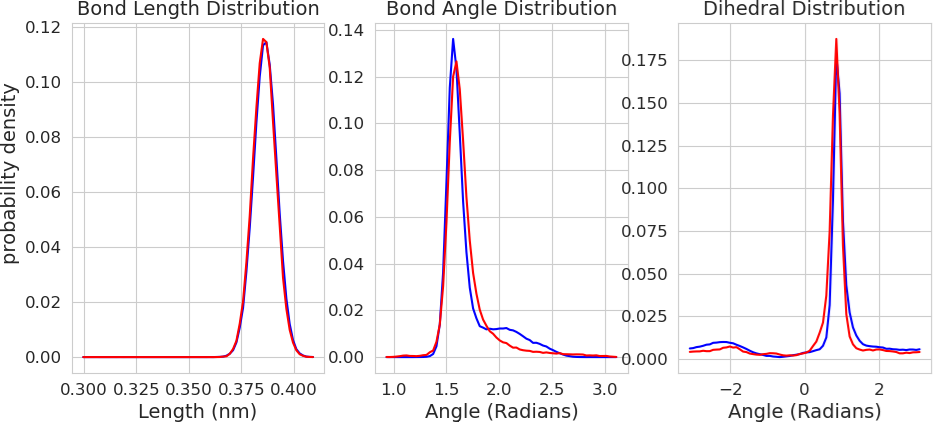}}\\
\subfloat[\textbf{FM+AA$_{\text{p}}$+Cov}]{\includegraphics[scale=0.5]{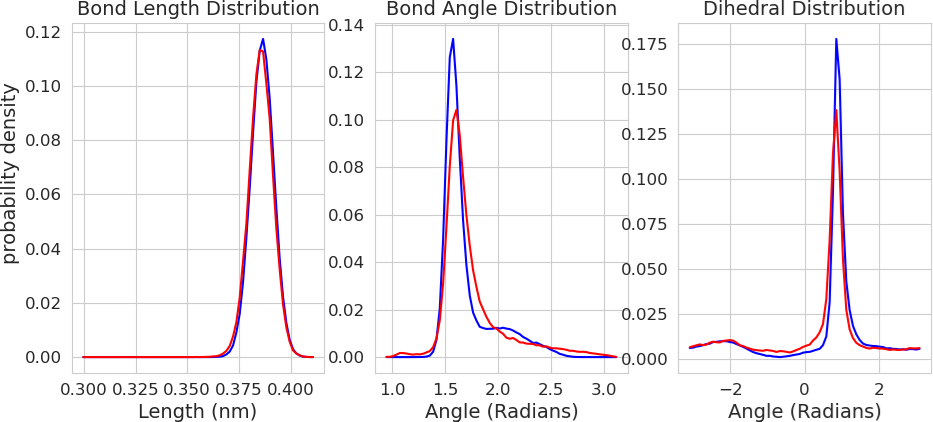}}
\caption{Structural distributions for $\alpha$3D showing bond lengths, bond angles, and dihedral angles. The W1 distances between each model distribution and ground-truth distribution are reported in Table \ref{tab:id_structural}.}
\label{fig:a3dStructure_id}
\end{figure}

We now show distributions for select proteins. For each protein and model, our pipeline generates eight relevant plots, three structural and five for slow-mode dynamics. Across all models and proteins, there are thus $8 \times 3 \times 9 = 216$ plots of interest. We thus show the plots for only two proteins---chignolin (Figures \ref{fig:chignolin2D_id}--\ref{fig:chignolinStructure_id}), which is the smallest, and $\alpha$3D (Figures \ref{fig:a3d2D_id}--\ref{fig:a3dStructure_id}), which shows the most complete improvement in slow-mode metrics from full HVP matching (including the covariance correction), despite being one of the largest proteins. We first show the TICA projections, including the Gaussian kernel density estimate along TIC0 and TIC1. We then show the structural distributions, specifically for bond length, bond angle, and dihedral angle.

\newpage

\subsection{Errors}

\begin{figure}[H]
    \centering
    \includegraphics[width=1\linewidth]{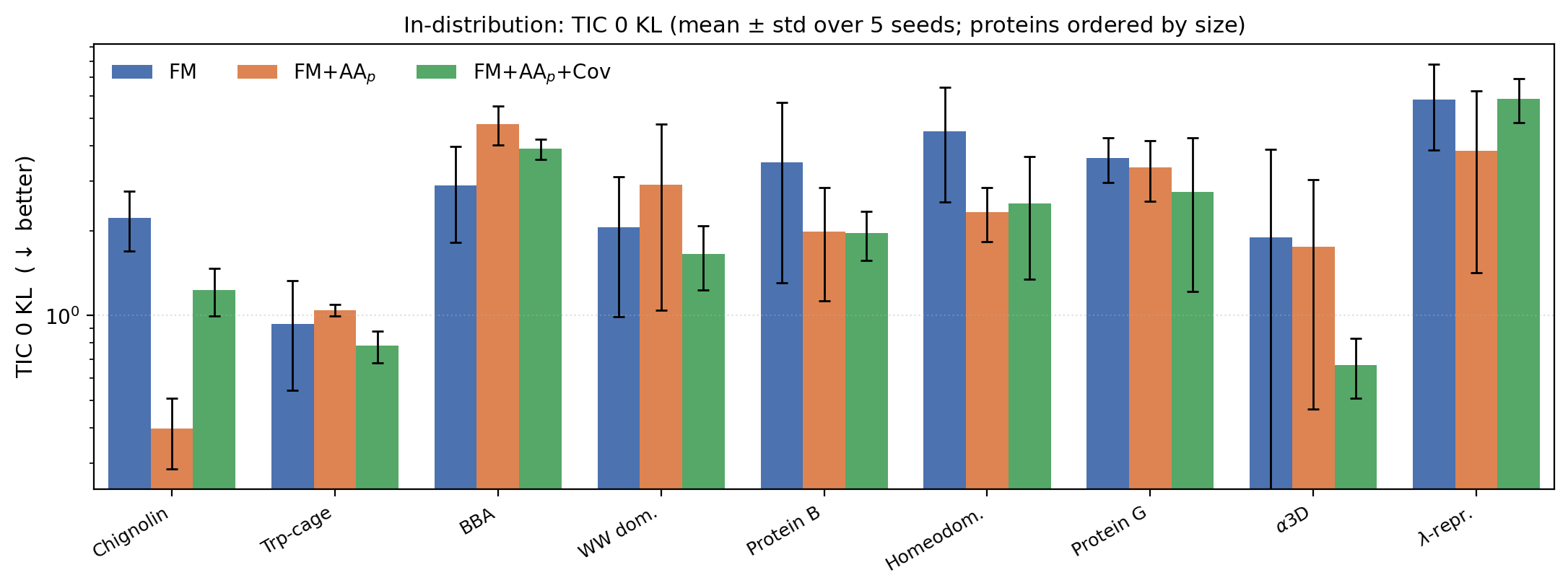}
    \caption{In-distribution TIC0 KL divergence. Grouped bars show the KL divergence along the slowest collective mode for each variant (FM, FM+AA$_\text{p}$, FM+AA$_\text{p}$+Cov), mean over 5 seeds with whiskers denoting one standard deviation, for models trained and evaluated on the nine benchmark proteins. Proteins are ordered by size (bead count) from left to right; lower is better. The $y$-axis is logarithmic.}

    \label{fig:id_tic0_kl_errors}
\end{figure}

\begin{figure}[H]
    \centering
    \includegraphics[width=1\linewidth]{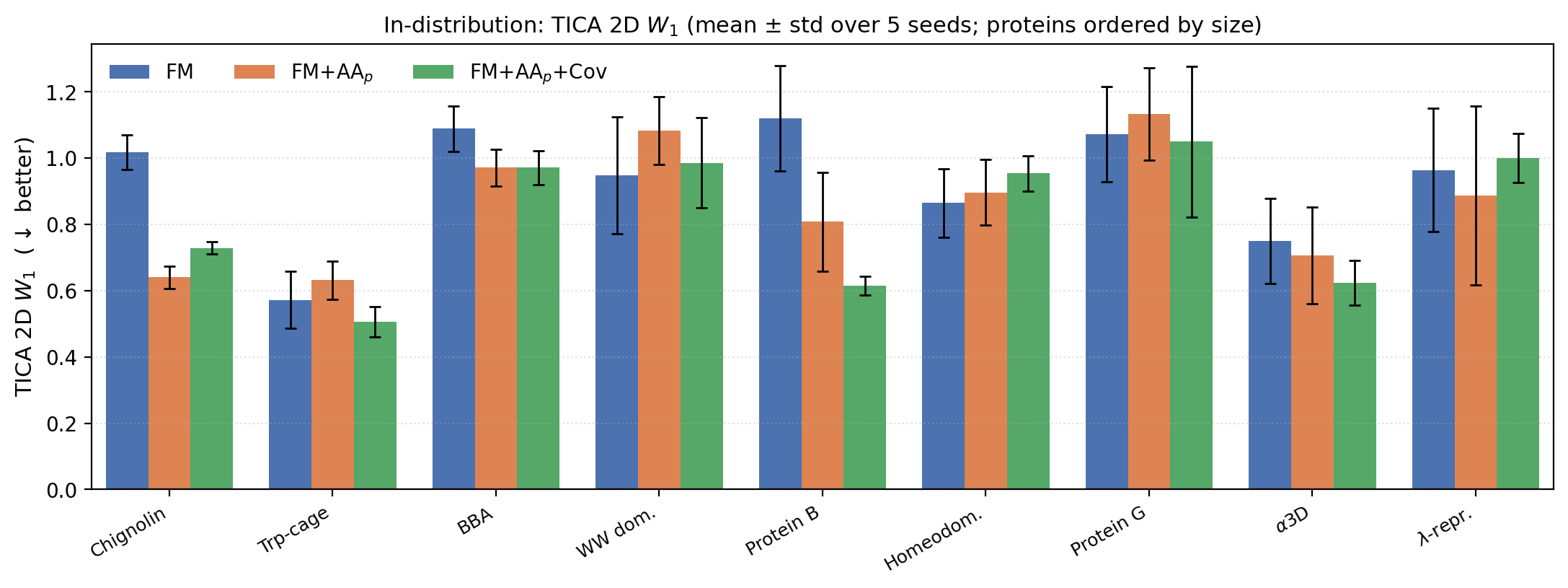}
    \caption{In-distribution TICA~2D W1 distance. Grouped bars show the W1 distance over the joint distribution of the two slowest components (TIC0 and TIC1) for each variant (FM, FM+AA$_\text{p}$, FM+AA$_\text{p}$+Cov), mean over 5 seeds with whiskers denoting one standard deviation, for models trained and evaluated on the nine benchmark proteins. Proteins are ordered by size (bead count) from left to right; lower is better.}
    \label{fig:id_tic2d_w1_errors}
\end{figure}

Another method of visualization for our metrics is to use box-and-whiskers plots. We show these below for the TIC0 KL divergence (Figure \ref{fig:id_tic0_kl_errors}) and TICA 2D W1 distance (Figure \ref{fig:id_tic2d_w1_errors}).

\newpage

\section{Generalization to unseen proteins}\label{app:generalization}

We now discuss an ablation experiment testing the effects of HVP matching on simulating the dynamics of proteins that are not directly included in the training set (out-of-distribution modeling). This is a more uncontrolled experiment than the in-distribution one of the main text, and so the results here should be viewed as preliminary. Nonetheless, the findings here are suggestive of the utility of HVP matching in out-of-distribution settings.

We use the same model variants as those listed in the main text (Section \ref{sec:experiments}). However, to train, we use a curated dataset of 99 single-chain proteins, constructed to include at least 2 proteins containing each bond type, 4 reference proteins (6MRR, 2VQ4, 5YM7, 3ID4), and additional proteins to bring the total to 99. The dataset has been filtered to exclude short (cis conformation) bonds; disulfide bonds were excluded at the base dataset level. The full set contains 15,522 C$_{\alpha}$ atoms:
\begin{quote}
\texttt{1AAJ 1ACF 1AMM 1AYD 1BK2 1BZ4 1CM2 1DVN 1EQ6 1EYH 1FNA 1GYV 1I2T 1IO2 1J7X 1JMW 1JW4 1K40 1LPJ 1MJC 1N81 1NG6 1OGW 1P1L 1QTP 1QZM 1UKF 1W8V 1WP5 1WWI 1WY3 1X3O 1XGD 1XT0 1YP5 1YU5 1YW5 1ZEQ 1ZLM 2BFH 2EVB 2H0M 2ICT 2NSC 2OJ4 2RH3 2V14 2VQ4 2W9Q 3HVM 3ID4 3KJE 3LFO 3MX7 3MZZ 3P7K 3PG4 3R6D 3RJP 4BPF 4J5Q 4N6T 4QAJ 4QBE 4RZ9 4U3H 4Y2K 5BTH 5GY3 5IHW 5J2V 5UVR 5XEF 5Y8E 5YM7 6APK 6DR3 6GPM 6KND 6L7Q 6LG3 6MRR 6Q3V 6SYG 6TYY 6WEY 6WIN 7DMF 7DMS 7LIQ 7TGP 7XCD 8A8S 8ERE 8G22 8GQQ 8HQW 8I2D 8JED}
\end{quote}
We train on 10{,}000 frames per protein (${\sim}990{,}000$ total training frames). The benchmark set~\cite{Aghili2025} is then used purely to evaluate the three model variants.

As in Appendix \ref{app:inDistSupplement}, we present the full set of metrics for both the slow-mode dynamics (KL divergences for the TIC0--3 projections and W1 distance for TICA 2D, Table~\ref{tab:ood_slow}) and the structural distributions (W1 distances for bond-length, bond-angle, radius-of-gyration, and dihedral distributions, Table~\ref{tab:ood_structural}). We also report training and validation loss curves (Figures~\ref{fig:train_loss_ood}--\ref{fig:val_loss_ood}) and learned density functions for chignolin (Figures~\ref{fig:chignolin2D_ood}--\ref{fig:chignolinStructure_ood}) and $\alpha$3D (Figures~\ref{fig:a3d2D_ood}--\ref{fig:a3dStructure_ood}).

\subsection{Slow-mode metrics}
\begin{figure}
\centering
\subfloat[\textbf{FM}]{\includegraphics[scale=0.45]{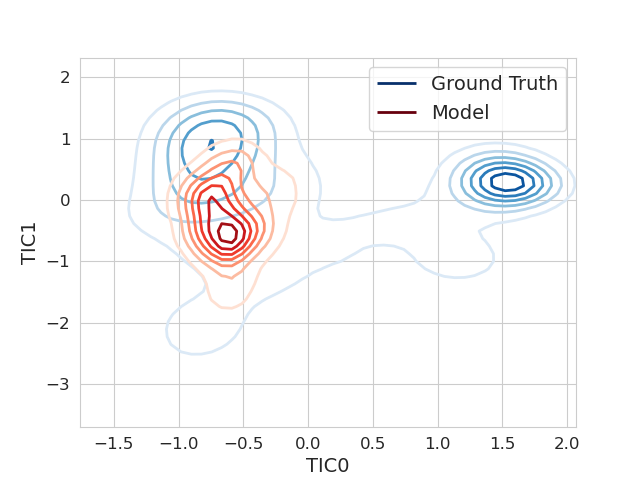}}
\subfloat[\textbf{FM+AA$_{\text{p}}$}]{\includegraphics[scale=0.45]{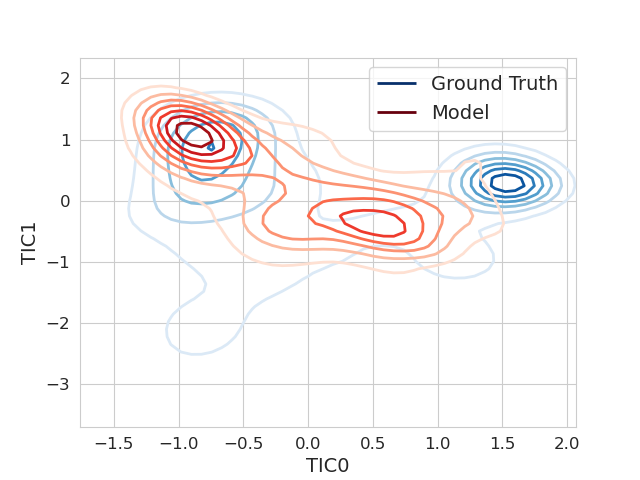}}
\caption{TICA free-energy contours for chignolin. Blue contours show the ground truth (AA reference), and red contours show the CG model. (a) \textbf{FM:}~the model's density concentrates in a single basin ($\text{TIC0}\approx-0.5$) with no density in the secondary basin ($\text{TIC0} \approx 1.5$). (b) \textbf{FM+AA$_\text{p}$:}~the model reproduces both basins more accurately with continuous density bridging the transition region, demonstrating that the curvature signal from HVP matching enables meaningful extrapolation to protein conformations not seen during training.}
\label{fig:tica_contours_chignolin}
\end{figure}

The slow-mode metrics are presented in Table~\ref{tab:ood_slow}. An HVP-matching variant achieves the best TICA~2D Wasserstein distance on seven of nine proteins and the best TIC0 KL divergence on seven of nine, indicating that the curvature signal learned from the training proteins can improve the slow-mode distributions of unseen sequences. However, whether just Term~1 (the projection of the all-atom Hessian) or that plus Term~2 (the covariance correction) performs better depends on the system.

Term~1 alone is sufficient for the smallest system, chignolin. For this protein, FM+AA$_\text{p}$ is the best variant on every slow-mode metric, reducing TICA~2D~W1 by 44\% and TIC0~KL by 80\%. Figure~\ref{fig:tica_contours_chignolin} illustrates this improvement qualitatively, where FM+AA$_\text{p}$ recovers both major basins and the density between them more accurately than force matching. In contrast, adding the covariance correction is substantially less effective (e.g.\ TIC2~KL increases from 0.30 under FM+AA$_\text{p}$ to 3.26). This suggests that the model-dependent correction in Term~2 can introduce error when applied to a system for which the projected Hessian term alone already provides an effective curvature signal.

However, the covariance correction becomes increasingly important for larger systems. For most proteins with 20 or more CG beads, FM+AA$_\text{p}$+Cov produces the best joint slow-mode accuracy. It achieves the best TICA~2D~W1 on Trp-cage, WW domain, Protein~B, Protein~G, $\alpha$3D, and $\lambda$-repressor, with reductions relative to force matching of 24\%, 13\%, 26\%, 30\%, and 26\% on the first five systems, respectively. In fact, Term~1 without the covariance correction degrades slow-mode accuracy on the larger proteins. FM+AA$_{\text{p}}$ performs worse than plain force matching on all but one of the TICA metrics for homeodomain, Protein G, $\alpha$3D, and $\lambda$-repressor, with TIC0~KL alone increasing by 140\%, 27\%, 125\%, and 114\%, respectively.

$\alpha$3D in particular highlights the complementary roles of the two HVP terms. The projected-Hessian term alone substantially degrades performance, but the full objective reverses these degradations. FM+AA$_\text{p}$+Cov achieves a TICA~2D~W1 of 0.888, a 26\% reduction relative to force matching, and a TIC0~KL of 1.30, a 67\% reduction. Thus, the covariance correction is crucial in the observed improvement to slow-mode accuracy. The contrast with chignolin suggests that the relative utility of the two curvature terms depends on the target system and its relationship to the training distribution, rather than on system size alone.

\newpage


\begin{table}[h]
\centering
\setlength{\tabcolsep}{6pt}
\begin{tabular}{@{}l@{\hspace{6pt}}lrrrrr@{}}
\toprule
\multicolumn{7}{c}{\textbf{Slow-mode metrics}} \\
\midrule
Protein & Model & \makecell{TICA 2D\\W1 $\downarrow$} & \makecell{TIC 0\\KL $\downarrow$} & \makecell{TIC 1\\KL $\downarrow$} & \makecell{TIC 2\\KL $\downarrow$} & \makecell{TIC 3\\KL $\downarrow$} \\
\midrule
\multirow{3}{*}{\shortstack[l]{Chignolin\\ $N=10$}} & FM & $0.867_{\pm 0.032}$ & $2.77_{\pm 0.68}$ & $0.61_{\pm 0.09}$ & $1.85_{\pm 0.43}$ & $0.14_{\pm 0.03}$ \\
 & +AA$_{\text{p}}$ & $\mathbf{0.482_{\pm 0.021}}$ & $\mathbf{0.55_{\pm 0.16}}$ & $\mathbf{0.38_{\pm 0.11}}$ & $\mathbf{0.30_{\pm 0.03}}$ & $\mathbf{0.10_{\pm 0.03}}$ \\
 & +Cov & $0.668_{\pm 0.051}$ & $1.73_{\pm 1.02}$ & $0.74_{\pm 0.13}$ & $3.26_{\pm 0.40}$ & $0.78_{\pm 0.29}$ \\
\midrule
\multirow{3}{*}{\shortstack[l]{Trp-cage\\ $N=20$}} & FM & $1.436_{\pm 0.041}$ & $\mathbf{4.76_{\pm 1.39}}$ & $3.35_{\pm 1.11}$ & $\mathbf{0.86_{\pm 0.20}}$ & $1.00_{\pm 0.28}$ \\
 & +AA$_{\text{p}}$ & $1.153_{\pm 0.020}$ & $5.03_{\pm 0.67}$ & $3.78_{\pm 0.91}$ & $1.34_{\pm 0.41}$ & $1.01_{\pm 0.49}$ \\
 & +Cov & $\mathbf{1.093_{\pm 0.043}}$ & $4.78_{\pm 0.31}$ & $\mathbf{3.05_{\pm 0.45}}$ & $0.93_{\pm 0.37}$ & $\mathbf{0.79_{\pm 0.11}}$ \\
\midrule
\multirow{3}{*}{\shortstack[l]{BBA\\ $N=28$}} & FM & $\mathbf{1.194_{\pm 0.075}}$ & $\mathbf{3.87_{\pm 0.83}}$ & $\mathbf{0.84_{\pm 0.21}}$ & $1.34_{\pm 0.32}$ & $\mathbf{0.65_{\pm 0.17}}$ \\
 & +AA$_{\text{p}}$ & $1.725_{\pm 0.113}$ & $5.04_{\pm 0.54}$ & $2.74_{\pm 0.85}$ & $\mathbf{1.01_{\pm 0.18}}$ & $1.31_{\pm 0.30}$ \\
 & +Cov & $1.226_{\pm 0.019}$ & $5.03_{\pm 0.72}$ & $1.93_{\pm 0.59}$ & $1.19_{\pm 0.26}$ & $0.71_{\pm 0.24}$ \\
\midrule
\multirow{3}{*}{\shortstack[l]{WW domain\\ $N=34$}} & FM & $1.371_{\pm 0.040}$ & $8.37_{\pm 0.39}$ & $3.41_{\pm 0.28}$ & $3.39_{\pm 2.30}$ & $\mathbf{1.28_{\pm 0.42}}$ \\
 & +AA$_{\text{p}}$ & $1.308_{\pm 0.088}$ & $7.44_{\pm 0.62}$ & $2.99_{\pm 0.35}$ & $2.16_{\pm 0.47}$ & $1.44_{\pm 0.27}$ \\
 & +Cov & $\mathbf{1.196_{\pm 0.082}}$ & $\mathbf{6.83_{\pm 0.57}}$ & $\mathbf{2.87_{\pm 0.28}}$ & $\mathbf{2.15_{\pm 0.79}}$ & $1.50_{\pm 0.28}$ \\
\midrule
\multirow{3}{*}{\shortstack[l]{Protein B\\ $N=47$}} & FM & $1.039_{\pm 0.059}$ & $4.94_{\pm 0.96}$ & $2.34_{\pm 0.69}$ & $3.78_{\pm 2.61}$ & $2.23_{\pm 0.56}$ \\
 & +AA$_{\text{p}}$ & $1.135_{\pm 0.066}$ & $4.04_{\pm 0.90}$ & $2.47_{\pm 0.29}$ & $\mathbf{1.55_{\pm 0.27}}$ & $2.03_{\pm 0.30}$ \\
 & +Cov & $\mathbf{0.766_{\pm 0.130}}$ & $\mathbf{1.15_{\pm 0.47}}$ & $\mathbf{1.28_{\pm 0.29}}$ & $1.61_{\pm 0.24}$ & $\mathbf{1.48_{\pm 0.37}}$ \\
\midrule
\multirow{3}{*}{\shortstack[l]{Homeodomain\\ $N=54$}} & FM & $\mathbf{0.945_{\pm 0.097}}$ & $1.34_{\pm 0.38}$ & $\mathbf{0.74_{\pm 0.11}}$ & $1.39_{\pm 0.57}$ & $0.52_{\pm 0.33}$ \\
 & +AA$_{\text{p}}$ & $1.515_{\pm 0.034}$ & $3.21_{\pm 0.51}$ & $6.55_{\pm 0.34}$ & $6.05_{\pm 0.95}$ & $2.65_{\pm 1.39}$ \\
 & +Cov & $1.047_{\pm 0.108}$ & $\mathbf{1.08_{\pm 0.45}}$ & $1.20_{\pm 0.59}$ & $\mathbf{1.07_{\pm 0.25}}$ & $\mathbf{0.38_{\pm 0.15}}$ \\
\midrule
\multirow{3}{*}{\shortstack[l]{Protein G\\ $N=56$}} & FM & $1.184_{\pm 0.134}$ & $4.52_{\pm 2.11}$ & $1.74_{\pm 0.55}$ & $0.99_{\pm 0.36}$ & $1.22_{\pm 0.64}$ \\
 & +AA$_{\text{p}}$ & $1.289_{\pm 0.087}$ & $5.75_{\pm 1.27}$ & $2.66_{\pm 0.32}$ & $0.99_{\pm 0.08}$ & $2.25_{\pm 0.43}$ \\
 & +Cov & $\mathbf{0.832_{\pm 0.155}}$ & $\mathbf{1.66_{\pm 0.98}}$ & $\mathbf{0.99_{\pm 0.17}}$ & $\mathbf{0.70_{\pm 0.12}}$ & $\mathbf{0.64_{\pm 0.28}}$ \\
\midrule
\multirow{3}{*}{\shortstack[l]{$\alpha$3D\\ $N=73$}} & FM & $1.207_{\pm 0.141}$ & $3.97_{\pm 2.48}$ & $1.55_{\pm 0.23}$ & $0.88_{\pm 0.17}$ & $\mathbf{0.64_{\pm 0.09}}$ \\
 & +AA$_{\text{p}}$ & $1.463_{\pm 0.082}$ & $8.95_{\pm 0.27}$ & $2.24_{\pm 1.14}$ & $1.49_{\pm 0.34}$ & $1.86_{\pm 0.41}$ \\
 & +Cov & $\mathbf{0.888_{\pm 0.116}}$ & $\mathbf{1.30_{\pm 0.17}}$ & $\mathbf{1.20_{\pm 0.34}}$ & $\mathbf{0.70_{\pm 0.25}}$ & $0.91_{\pm 0.12}$ \\
\midrule
\multirow{3}{*}{\shortstack[l]{$\lambda$-repressor\\ $N=80$}} & FM & $0.783_{\pm 0.111}$ & $1.81_{\pm 0.86}$ & $0.89_{\pm 0.52}$ & $\mathbf{0.85_{\pm 0.27}}$ & $\mathbf{0.58_{\pm 0.19}}$ \\
 & +AA$_{\text{p}}$ & $0.977_{\pm 0.068}$ & $3.87_{\pm 1.98}$ & $1.69_{\pm 0.32}$ & $1.24_{\pm 0.12}$ & $1.55_{\pm 0.40}$ \\
 & +Cov & $\mathbf{0.772_{\pm 0.042}}$ & $\mathbf{1.74_{\pm 0.63}}$ & $\mathbf{0.86_{\pm 0.29}}$ & $1.10_{\pm 0.43}$ & $0.87_{\pm 0.44}$ \\
\bottomrule
\end{tabular}
\caption{Out-of-distribution benchmark, consisting of W1 distance on estimated density over TIC0--1 and KL divergences on TIC0--3. Metrics are computed over 5 seeds (mean$_{\pm\text{std}}$), with best per protein in bold; lower mean is better.}
\label{tab:ood_slow}
\end{table}

\newpage

\subsection{Structural metrics}

Local structural properties (bond lengths, angles, radius of gyration, and dihedrals) show a more mixed pattern. FM+AA$_\text{p}$ tends to improve bond-length distributions, while angle and dihedral improvements are protein-dependent and do not follow a consistent trend across model variants. These observations are expected, since local geometric properties are already directly constrained by the force matching objective. This contrasts with the more consistent improvements observed for the slow-mode distributions, suggesting that the primary benefit of curvature supervision in the OOD setting is not a uniform improvement in local structural statistics but rather improved representation of slow collective conformational behavior.

\begin{table}[H]

\centering
\setlength{\tabcolsep}{6pt}

\begin{tabular}{@{}l@{\hspace{6pt}}lrrrr@{}}
\toprule
\multicolumn{6}{c}{\textbf{Structural metrics (W1)}} \\
\midrule
Protein & Model & \makecell{Length\\$(\times 10^{-2})$ $\downarrow$} & Angle $\downarrow$ & RoG $\downarrow$ & Dihedral $\downarrow$ \\
\midrule
\multirow{3}{*}{\shortstack[l]{Chignolin\\ $N=10$}} & FM & $0.04_{\pm 0.01}$ & $0.082_{\pm 0.012}$ & $0.127_{\pm 0.002}$ & $0.627_{\pm 0.066}$ \\
 & +AA$_{\text{p}}$ & $\mathbf{0.02_{\pm 0.01}}$ & $\mathbf{0.059_{\pm 0.014}}$ & $\mathbf{0.090_{\pm 0.005}}$ & $\mathbf{0.354_{\pm 0.042}}$ \\
 & +Cov & $0.16_{\pm 0.03}$ & $0.218_{\pm 0.015}$ & $0.177_{\pm 0.002}$ & $0.644_{\pm 0.044}$ \\
\midrule
\multirow{3}{*}{\shortstack[l]{Trp-cage\\ $N=20$}} & FM & $0.11_{\pm 0.01}$ & $0.135_{\pm 0.008}$ & $0.201_{\pm 0.014}$ & $\mathbf{0.165_{\pm 0.015}}$ \\
 & +AA$_{\text{p}}$ & $\mathbf{0.11_{\pm 0.02}}$ & $\mathbf{0.062_{\pm 0.009}}$ & $0.122_{\pm 0.025}$ & $0.172_{\pm 0.019}$ \\
 & +Cov & $0.12_{\pm 0.02}$ & $0.120_{\pm 0.037}$ & $\mathbf{0.055_{\pm 0.010}}$ & $0.206_{\pm 0.006}$ \\
\midrule
\multirow{3}{*}{\shortstack[l]{BBA\\ $N=28$}} & FM & $0.24_{\pm 0.03}$ & $0.072_{\pm 0.012}$ & $0.132_{\pm 0.019}$ & $\mathbf{0.189_{\pm 0.037}}$ \\
 & +AA$_{\text{p}}$ & $0.18_{\pm 0.02}$ & $\mathbf{0.062_{\pm 0.008}}$ & $0.275_{\pm 0.021}$ & $0.445_{\pm 0.128}$ \\
 & +Cov & $\mathbf{0.08_{\pm 0.02}}$ & $0.094_{\pm 0.014}$ & $\mathbf{0.111_{\pm 0.023}}$ & $0.259_{\pm 0.083}$ \\
\midrule
\multirow{3}{*}{\shortstack[l]{WW domain\\ $N=34$}} & FM & $0.28_{\pm 0.01}$ & $\mathbf{0.064_{\pm 0.008}}$ & $\mathbf{0.053_{\pm 0.009}}$ & $\mathbf{0.476_{\pm 0.034}}$ \\
 & +AA$_{\text{p}}$ & $\mathbf{0.08_{\pm 0.02}}$ & $0.122_{\pm 0.009}$ & $0.239_{\pm 0.014}$ & $0.623_{\pm 0.062}$ \\
 & +Cov & $0.20_{\pm 0.05}$ & $0.075_{\pm 0.007}$ & $0.186_{\pm 0.026}$ & $0.522_{\pm 0.044}$ \\
\midrule
\multirow{3}{*}{\shortstack[l]{Protein B\\ $N=47$}} & FM & $0.09_{\pm 0.01}$ & $0.187_{\pm 0.033}$ & $0.186_{\pm 0.021}$ & $0.394_{\pm 0.027}$ \\
 & +AA$_{\text{p}}$ & $\mathbf{0.05_{\pm 0.01}}$ & $0.194_{\pm 0.017}$ & $\mathbf{0.094_{\pm 0.011}}$ & $0.462_{\pm 0.035}$ \\
 & +Cov & $0.45_{\pm 0.11}$ & $\mathbf{0.139_{\pm 0.032}}$ & $0.189_{\pm 0.024}$ & $\mathbf{0.354_{\pm 0.018}}$ \\
\midrule
\multirow{3}{*}{\shortstack[l]{Homeodomain\\ $N=54$}} & FM & $0.48_{\pm 0.09}$ & $\mathbf{0.083_{\pm 0.010}}$ & $0.101_{\pm 0.043}$ & $0.364_{\pm 0.032}$ \\
 & +AA$_{\text{p}}$ & $\mathbf{0.20_{\pm 0.02}}$ & $0.089_{\pm 0.008}$ & $0.119_{\pm 0.042}$ & $\mathbf{0.268_{\pm 0.014}}$ \\
 & +Cov & $0.25_{\pm 0.10}$ & $0.113_{\pm 0.016}$ & $\mathbf{0.092_{\pm 0.013}}$ & $0.333_{\pm 0.031}$ \\
\midrule
\multirow{3}{*}{\shortstack[l]{Protein G\\ $N=56$}} & FM & $0.96_{\pm 0.21}$ & $0.084_{\pm 0.008}$ & $0.213_{\pm 0.051}$ & $\mathbf{0.373_{\pm 0.040}}$ \\
 & +AA$_{\text{p}}$ & $0.31_{\pm 0.04}$ & $\mathbf{0.055_{\pm 0.006}}$ & $0.216_{\pm 0.042}$ & $0.429_{\pm 0.017}$ \\
 & +Cov & $\mathbf{0.27_{\pm 0.10}}$ & $0.066_{\pm 0.007}$ & $\mathbf{0.160_{\pm 0.024}}$ & $0.489_{\pm 0.101}$ \\
\midrule
\multirow{3}{*}{\shortstack[l]{$\alpha$3D\\ $N=73$}} & FM & $0.88_{\pm 0.16}$ & $0.142_{\pm 0.029}$ & $0.180_{\pm 0.028}$ & $\mathbf{0.449_{\pm 0.120}}$ \\
 & +AA$_{\text{p}}$ & $\mathbf{0.08_{\pm 0.01}}$ & $0.243_{\pm 0.010}$ & $0.232_{\pm 0.029}$ & $0.648_{\pm 0.066}$ \\
 & +Cov & $0.73_{\pm 0.33}$ & $\mathbf{0.136_{\pm 0.048}}$ & $\mathbf{0.153_{\pm 0.050}}$ & $0.465_{\pm 0.106}$ \\
\midrule
\multirow{3}{*}{\shortstack[l]{$\lambda$-repressor\\ $N=80$}} & FM & $0.38_{\pm 0.20}$ & $0.162_{\pm 0.022}$ & $0.107_{\pm 0.033}$ & $0.294_{\pm 0.036}$ \\
 & +AA$_{\text{p}}$ & $0.21_{\pm 0.06}$ & $0.134_{\pm 0.010}$ & $\mathbf{0.104_{\pm 0.025}}$ & $\mathbf{0.277_{\pm 0.026}}$ \\
 & +Cov & $\mathbf{0.17_{\pm 0.15}}$ & $\mathbf{0.131_{\pm 0.010}}$ & $0.156_{\pm 0.026}$ & $0.291_{\pm 0.032}$ \\
\bottomrule
\end{tabular}
\caption{Out-of-distribution benchmark, consisting of W1 distances on bond-length, bond-angle, radius-of-gyration, and dihedral distributions. Metrics are computed over 5 seeds (mean$_{\pm\text{std}}$), with best per protein in bold; lower mean is better.}
\label{tab:ood_structural}
\end{table}
\newpage

\subsection{Training and validation loss comparisons}

\begin{figure}[H]
 \centering
 \includegraphics[scale=0.25]{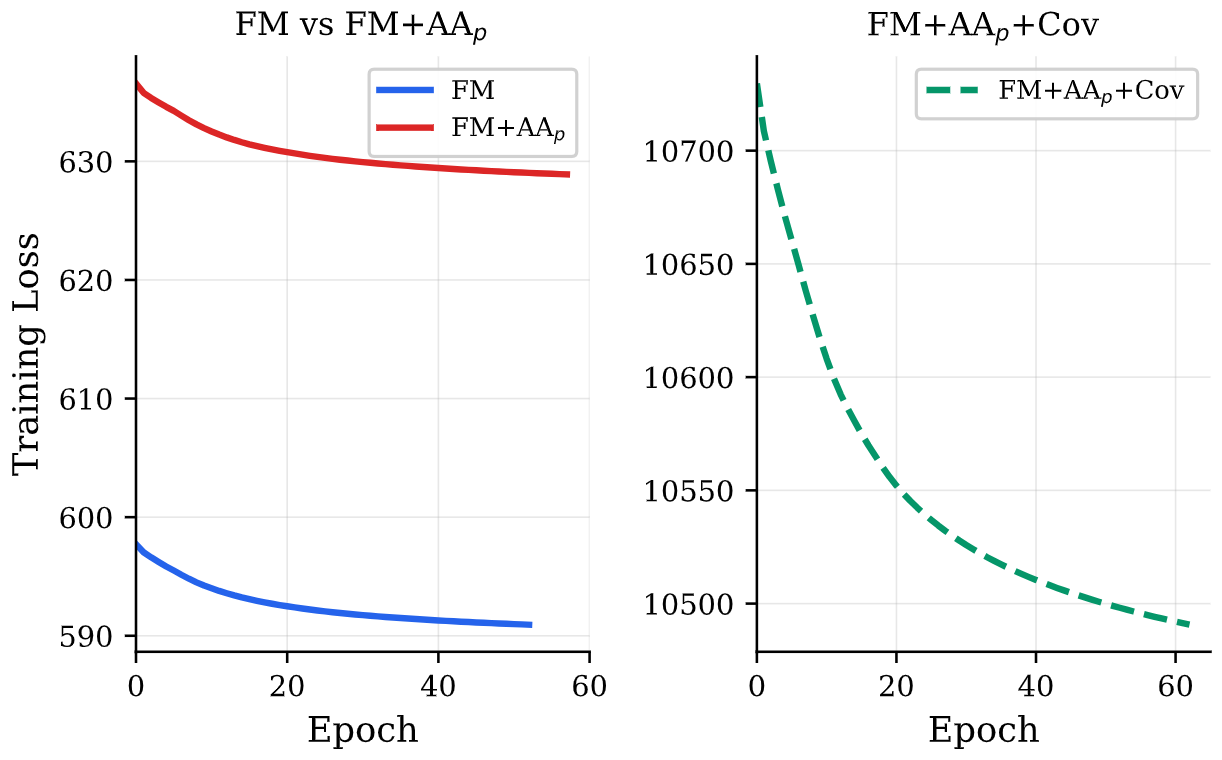}
 \caption{Training loss decomposition for the out-of-distribution models. \textbf{Left:}~FM and FM+AA$_\text{p}$ training losses, where the $\sim$38~kcal$^2$\,mol$^{-2}$\,\AA$^{-2}$ offset reflects the HVP matching term ($w_{\mathrm{HVP}} \|\mathbf{H}_{\mathrm{CG}}\mathbf{v} - \mathbf{H}_\theta\mathbf{v}\|^2$). \textbf{Right:}~ FM+AA$_\text{p}$+Cov training loss, dominated by the covariance correction (Term~2). Despite the order-of-magnitude difference in training-loss scales, all three models achieve equivalent force prediction accuracy on held-out data (Figure~\ref{fig:val_loss_ood}).}
 \label{fig:train_loss_ood}
 \end{figure}

\begin{figure}[H]
 \centering
 \includegraphics[scale=0.25]{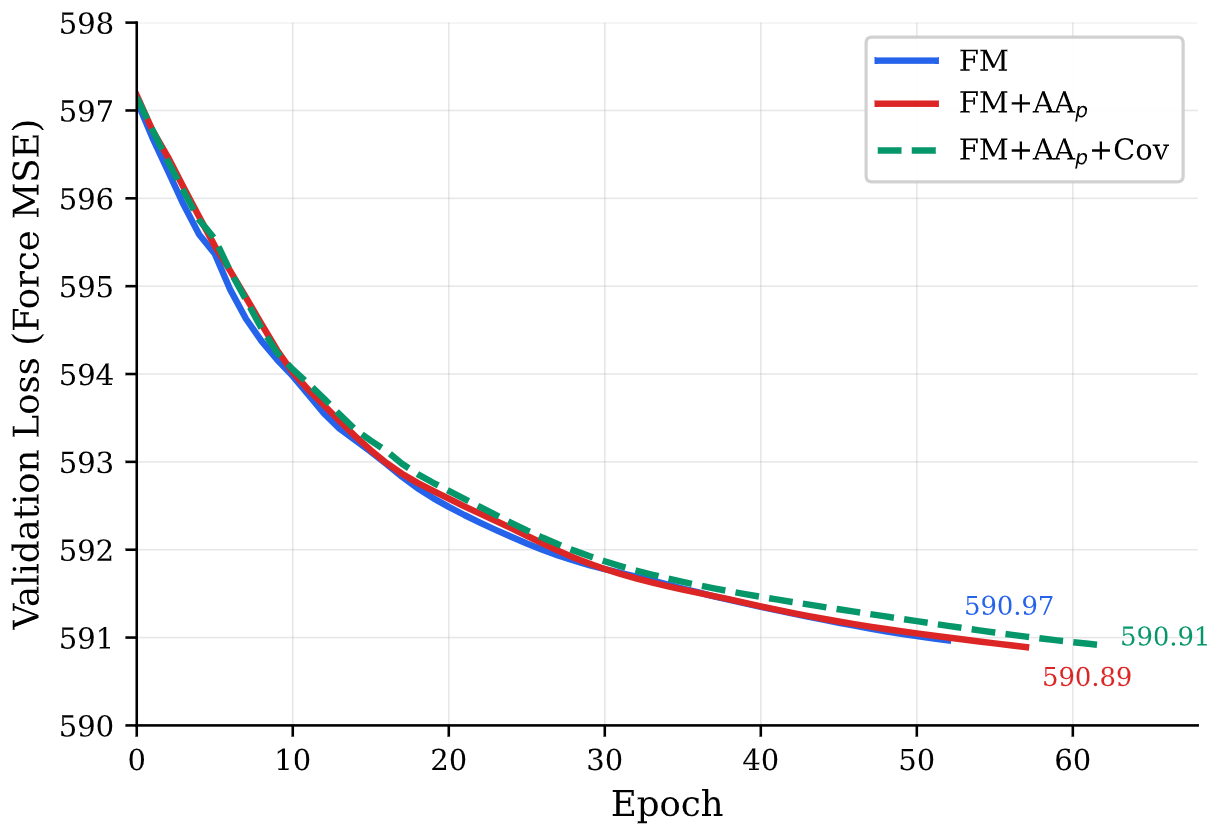}
 \caption{Validation loss (force matching MSE) during training on the single-chain dataset at $T=300$\,K. All three objectives (FM, FM+AA$_\text{p}$, and FM+AA$_\text{p}$+Cov) converge to comparable validation performance (${\sim}590.9$~kcal$^2$\,mol$^{-2}$\,\AA$^{-2}$), confirming that the auxiliary curvature-matching terms do not degrade force prediction accuracy. Models were trained for 52 (FM), 57 (FM+AA$_\text{p}$), and 62 (FM+AA$_\text{p}$+Cov) epochs with $w_\text{HVP} = 0.01$ and batch size 200 on 4$\times$A40 GPUs.}
\label{fig:val_loss_ood}
\end{figure}

We report loss curves for the out-of-distribution models, with Figure \ref{fig:train_loss_ood} visualizing convergence and Figure \ref{fig:val_loss_ood} visualizing force-prediction error.

\subsection{Distributions}

\begin{figure}[h]
\centering
\subfloat[\textbf{FM}]{\includegraphics[scale=0.45]{figures/ood_densities/chignolin_fin/FM/2d.png}}\\
\subfloat[\textbf{FM+AA$_{\text{p}}$}]{\includegraphics[scale=0.45]{figures/ood_densities/chignolin_fin/AAp/2d.png}}
\subfloat[\textbf{FM+AA$_{\text{p}}$+Cov}]{\includegraphics[scale=0.45]{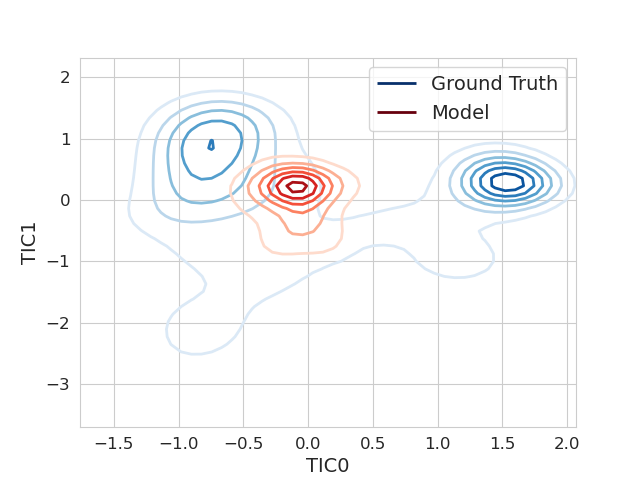}}
\caption{TICA 2D distributions for chignolin showing the Gaussian kernel density estimates of the free energy projected along TIC0 and TIC1. W1 distances between each model distribution and ground-truth distribution are reported in Table \ref{tab:ood_slow}. In addition to the distributions shown in Figure \ref{fig:tica_contours_chignolin}, we have also included the FM+AA$_{\text{p}}$+Cov distribution, which fails to learn the secondary basin.}
\label{fig:chignolin2D_ood}
\end{figure}

\begin{figure}
\centering
\subfloat[\textbf{FM}]{\includegraphics[scale=0.425]{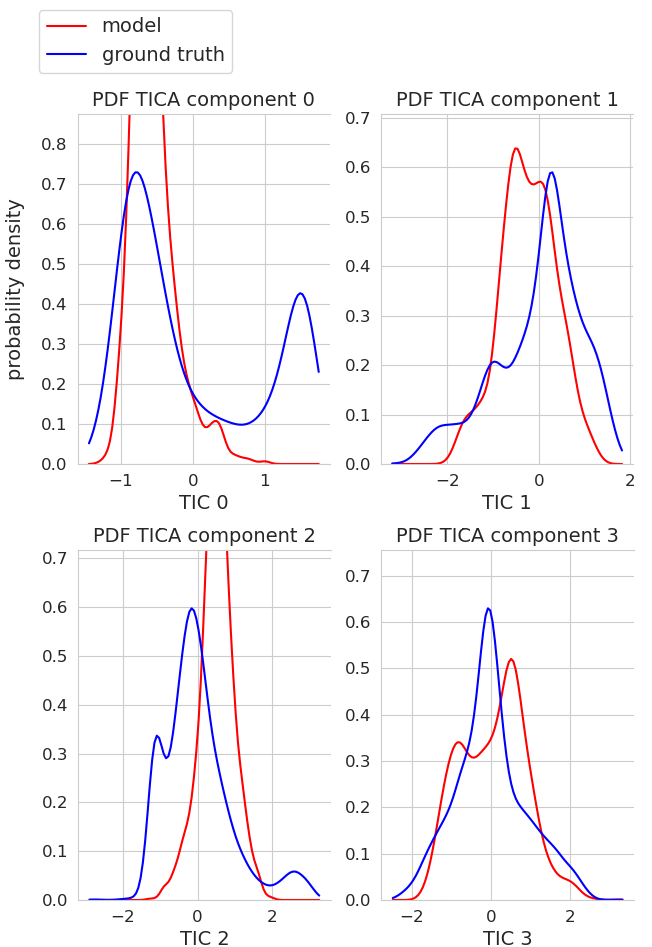}}\\
\subfloat[\textbf{FM+AA$_{\text{p}}$}]{\includegraphics[scale=0.425]{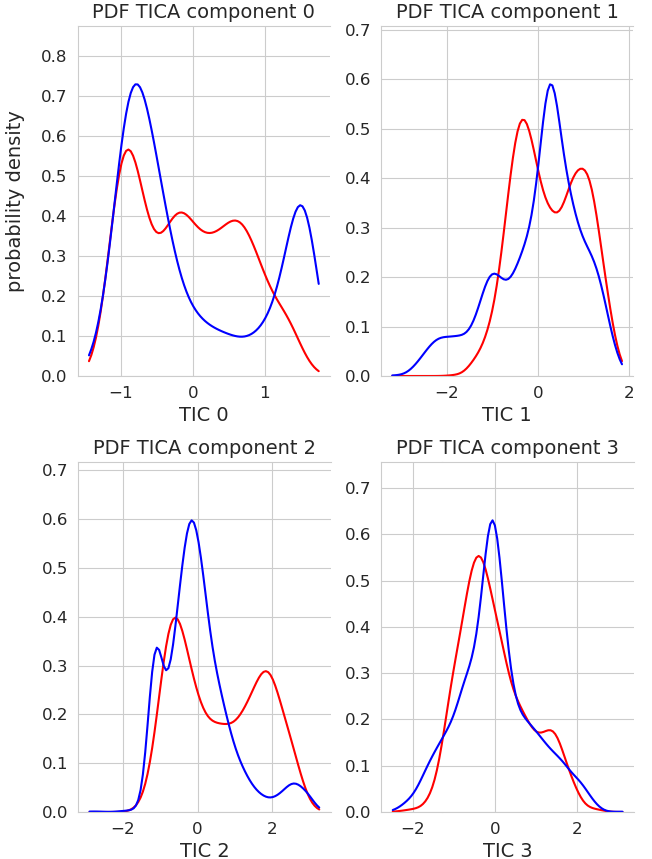}}
\subfloat[\textbf{FM+AA$_{\text{p}}$+Cov}]{\includegraphics[scale=0.425]{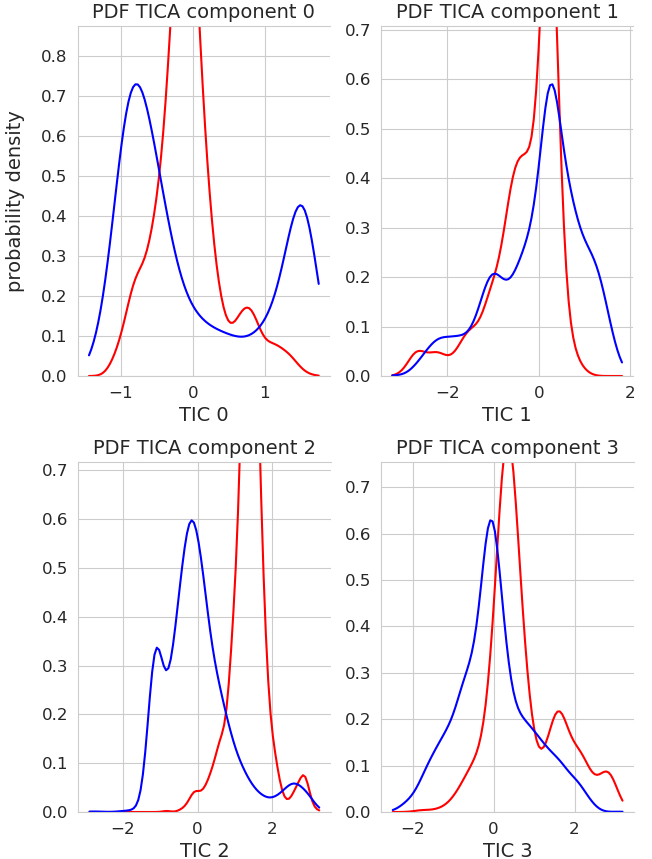}}
\caption{Slow-mode distributions for chignolin showing TIC0--3 projections of the free energy. The KL divergences between each model distribution and ground-truth distribution are reported in Table \ref{tab:ood_slow}.}
\label{fig:chignolinTICA_ood}
\end{figure}

\begin{figure}
\centering
\subfloat[\textbf{FM}]{\includegraphics[scale=0.5]{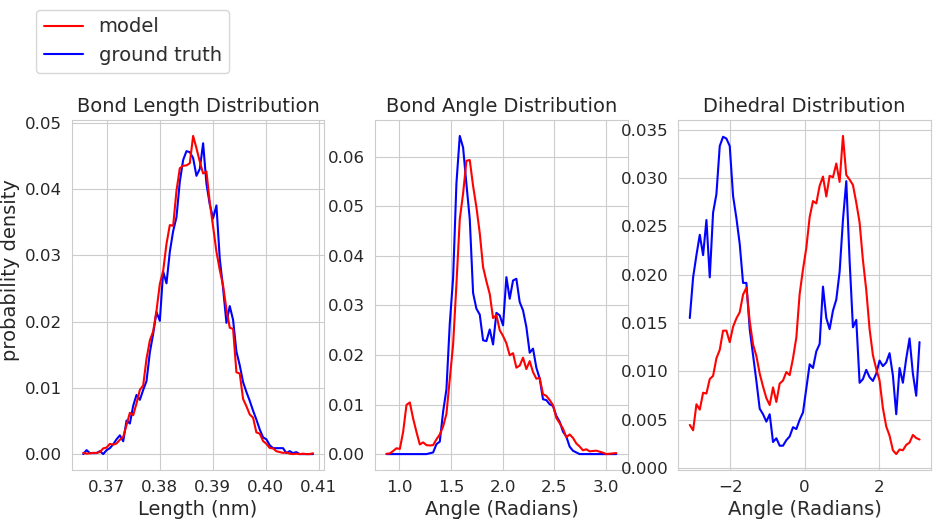}}\\
\subfloat[\textbf{FM+AA$_{\text{p}}$}]{\includegraphics[scale=0.5]{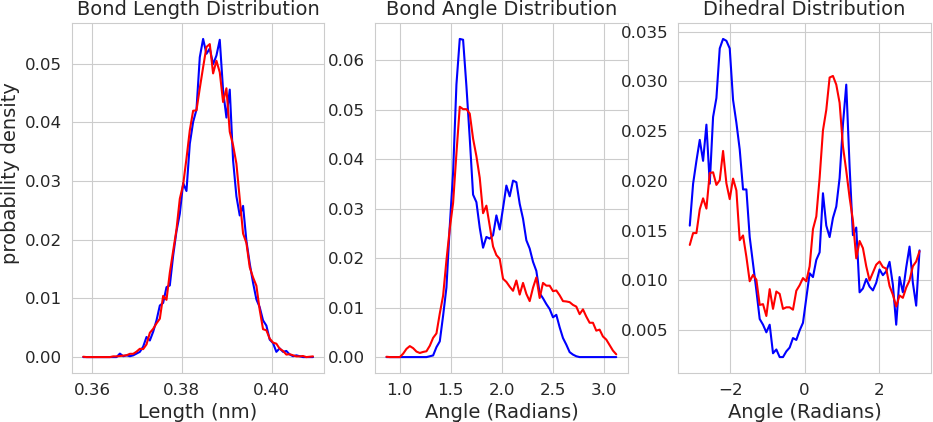}}\\
\subfloat[\textbf{FM+AA$_{\text{p}}$+Cov}]{\includegraphics[scale=0.5]{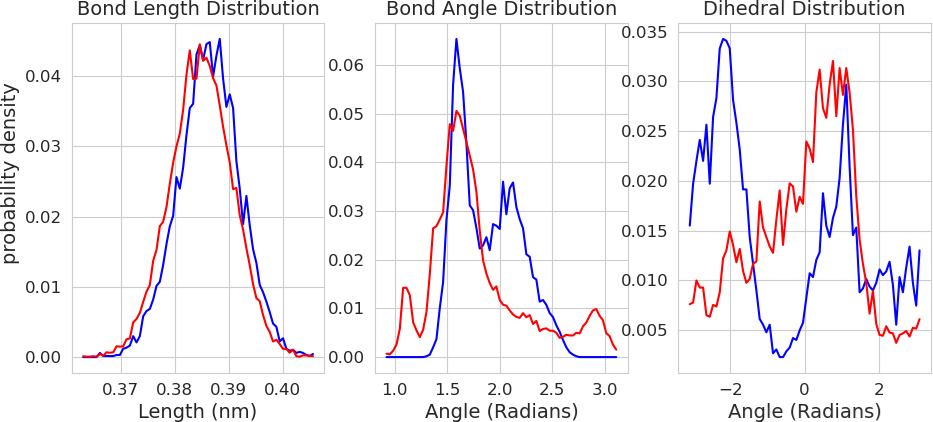}}
\caption{Structural distributions for chignolin showing bond lengths, bond angles, and dihedral angles. The W1 distances between each model distribution and ground-truth distribution are reported in Table \ref{tab:ood_structural}.}
\label{fig:chignolinStructure_ood}
\end{figure}

We now show distributions for chignolin (Figures \ref{fig:chignolin2D_ood}--\ref{fig:chignolinStructure_ood}) and $\alpha$3D (Figures \ref{fig:a3d2D_ood}--\ref{fig:a3dStructure_ood}), as in Appendix \ref{app:inDistSupplement}. In addition to consistency with our presentation of the extended in-distribution results, we choose these proteins because chignolin best reflects the efficacy of the FM+AA$_{\text{p}}$ model for small systems, particularly in reproducing TICA~2D, while $\alpha$3D shows well the importance of the covariance correction. For both proteins, we show the TICA projections, including the Gaussian kernel density estimate along TIC0 and TIC1, followed by the structural distributions for bond length, bond angle, and dihedral angle.

\begin{figure}
\centering
\subfloat[\textbf{FM}]{\includegraphics[scale=0.45]{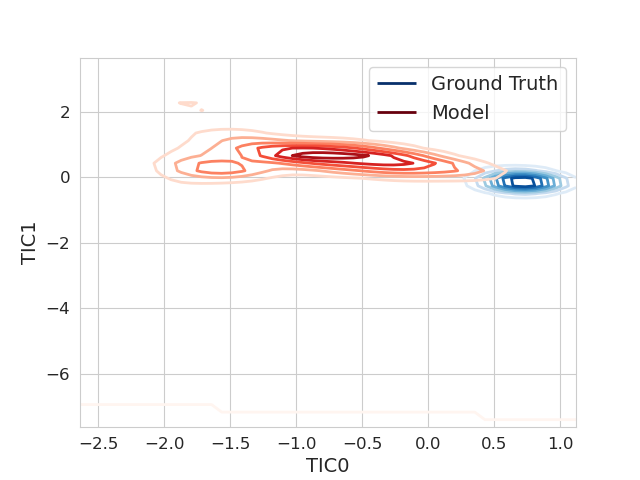}}\\
\subfloat[\textbf{FM+AA$_{\text{p}}$}]{\includegraphics[scale=0.45]{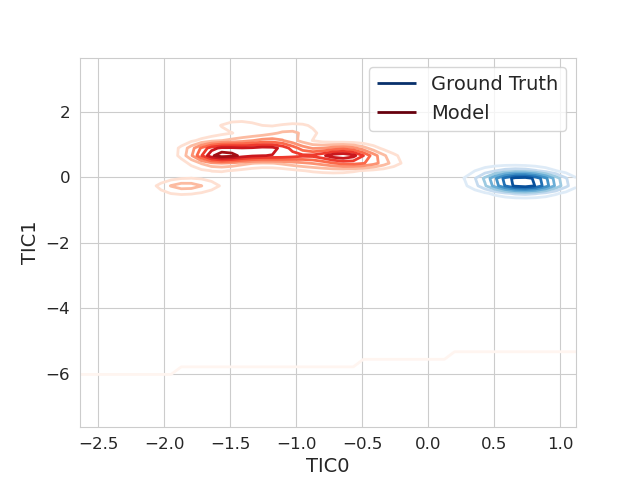}}
\subfloat[\textbf{FM+AA$_{\text{p}}$+Cov}]{\includegraphics[scale=0.45]{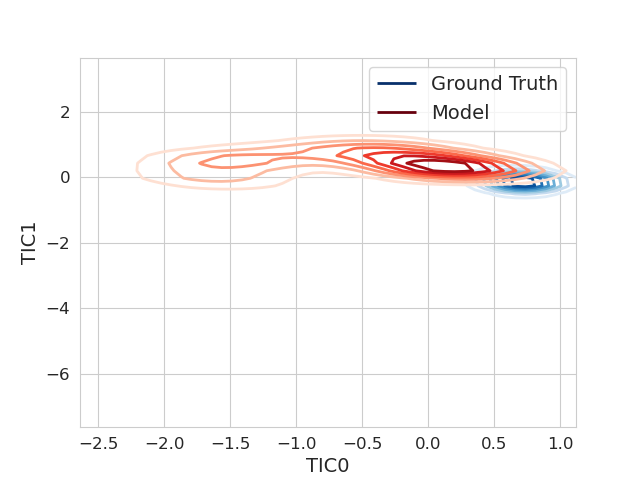}}
\caption{TICA 2D distributions for $\alpha$3D showing the Gaussian kernel density estimate of the free energy projected along TIC0 and TIC1. W1 distances between each model distribution and ground-truth distribution are reported in Table \ref{tab:ood_slow}.}
\label{fig:a3d2D_ood}
\end{figure}

\begin{figure}
\centering
\subfloat[\textbf{FM}]{\includegraphics[scale=0.425]{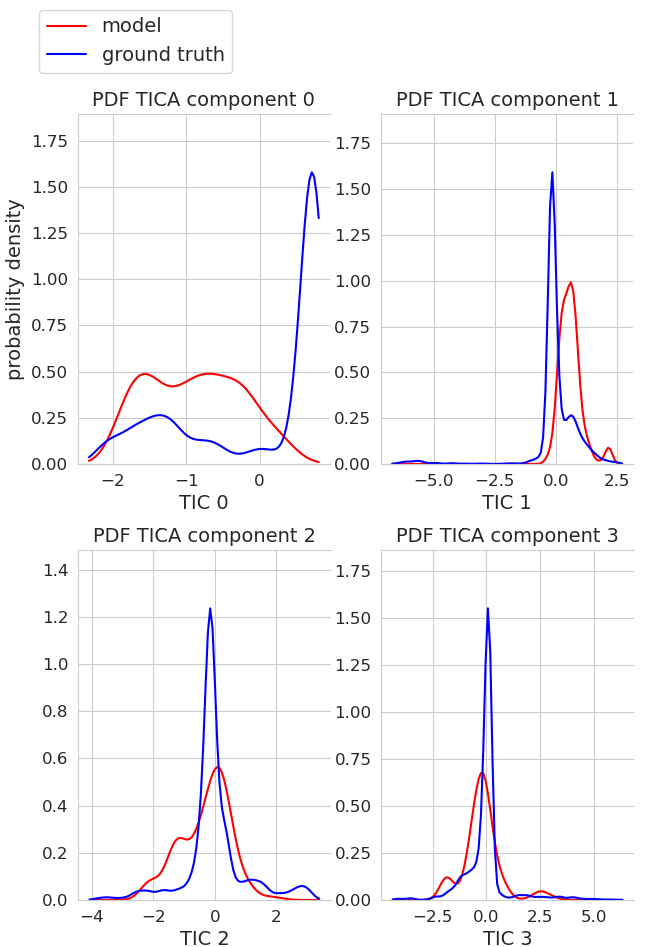}}\\
\subfloat[\textbf{FM+AA$_{\text{p}}$}]{\includegraphics[scale=0.425]{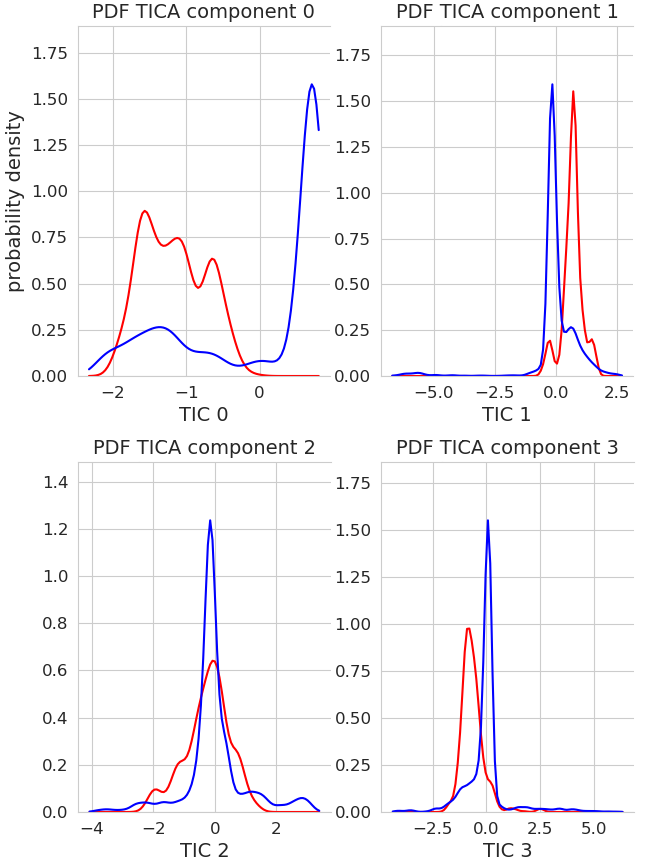}}
\subfloat[\textbf{FM+AA$_{\text{p}}$+Cov}]{\includegraphics[scale=0.425]{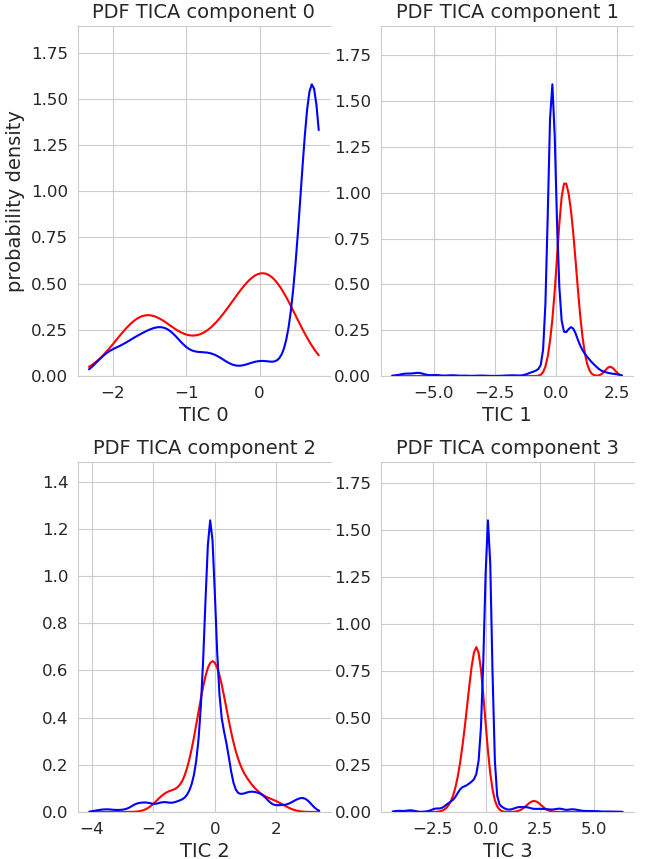}}
\caption{Slow-mode distributions for $\alpha$3D showing TIC0--3 projections of the free energy. The KL divergences between each model distribution and ground-truth distribution are reported in Table \ref{tab:ood_slow}.}
\label{fig:a3dTICA_ood}
\end{figure}

\begin{figure}
\centering
\subfloat[\textbf{FM}]{\includegraphics[scale=0.5]{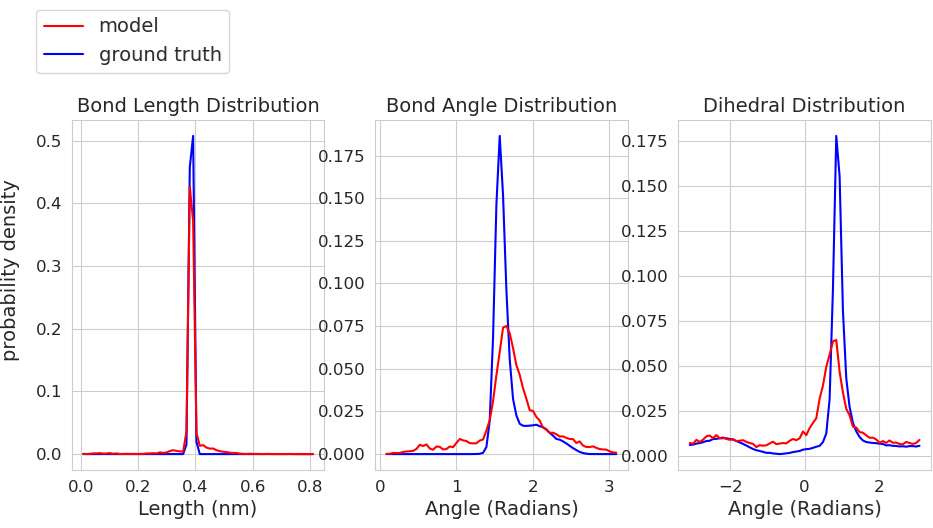}}\\
\subfloat[\textbf{FM+AA$_{\text{p}}$}]{\includegraphics[scale=0.5]{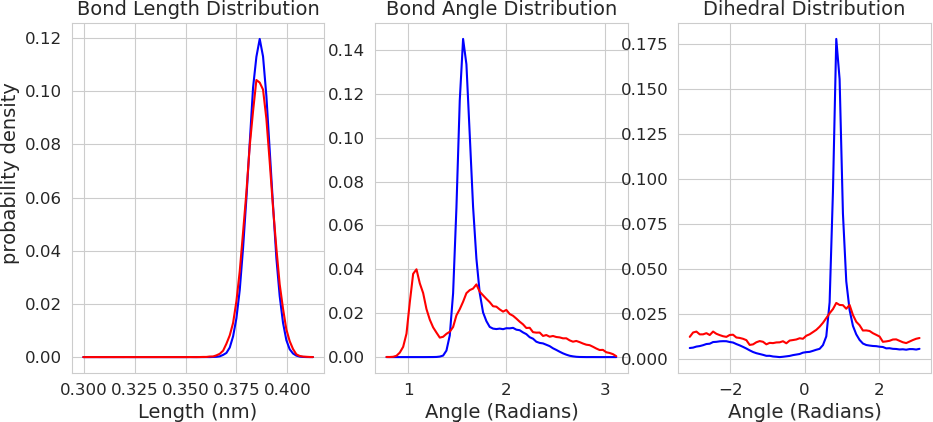}}\\
\subfloat[\textbf{FM+AA$_{\text{p}}$+Cov}]{\includegraphics[scale=0.5]{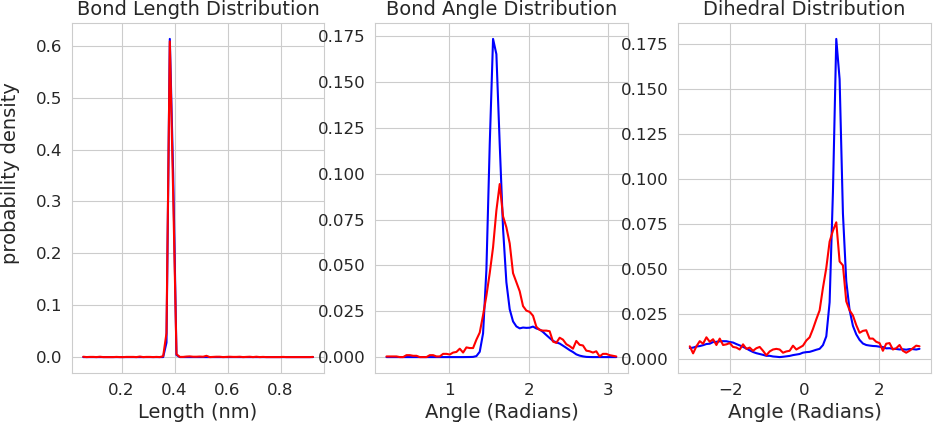}}
\caption{Structural distributions for $\alpha$3D showing bond lengths, bond angles, and dihedral angles. The W1 distances between each model distribution and ground-truth distribution are reported in Table \ref{tab:ood_structural}.}
\label{fig:a3dStructure_ood}
\end{figure}

\newpage

\subsection{Errors}

\begin{figure}[H]
    \centering
    \includegraphics[width=1\linewidth]{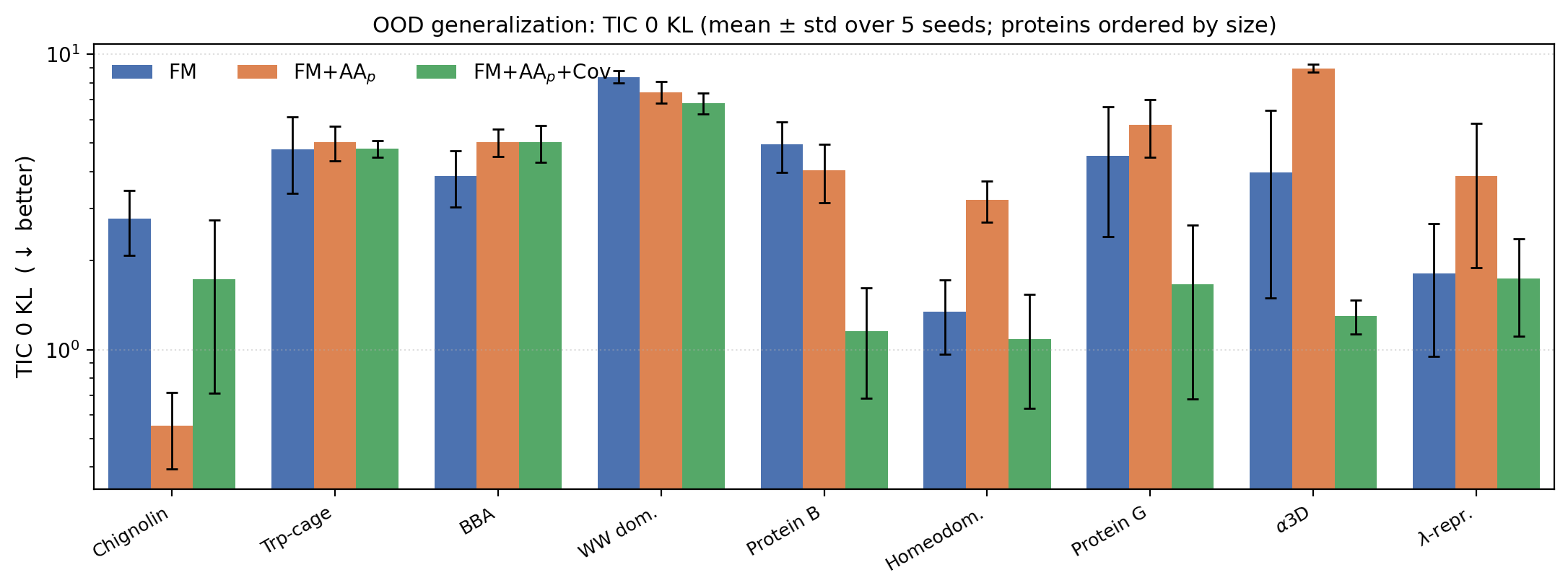}
    \caption{Out-of-distribution TIC0 KL divergence. Grouped bars show the KL divergence along the slowest collective mode for each variant (FM, FM+AA$_\text{p}$, FM+AA$_\text{p}$+Cov), mean over 5 seeds with whiskers denoting one standard deviation, for models trained on the single-chain dataset and evaluated on the nine benchmark proteins. Proteins are ordered by size (bead count) from left to right; lower is better. The $y$-axis is logarithmic.}
    \label{fig:ood_tic0_kl_errors}
\end{figure}

\begin{figure}[H]
    \centering
    \includegraphics[width=1\linewidth]{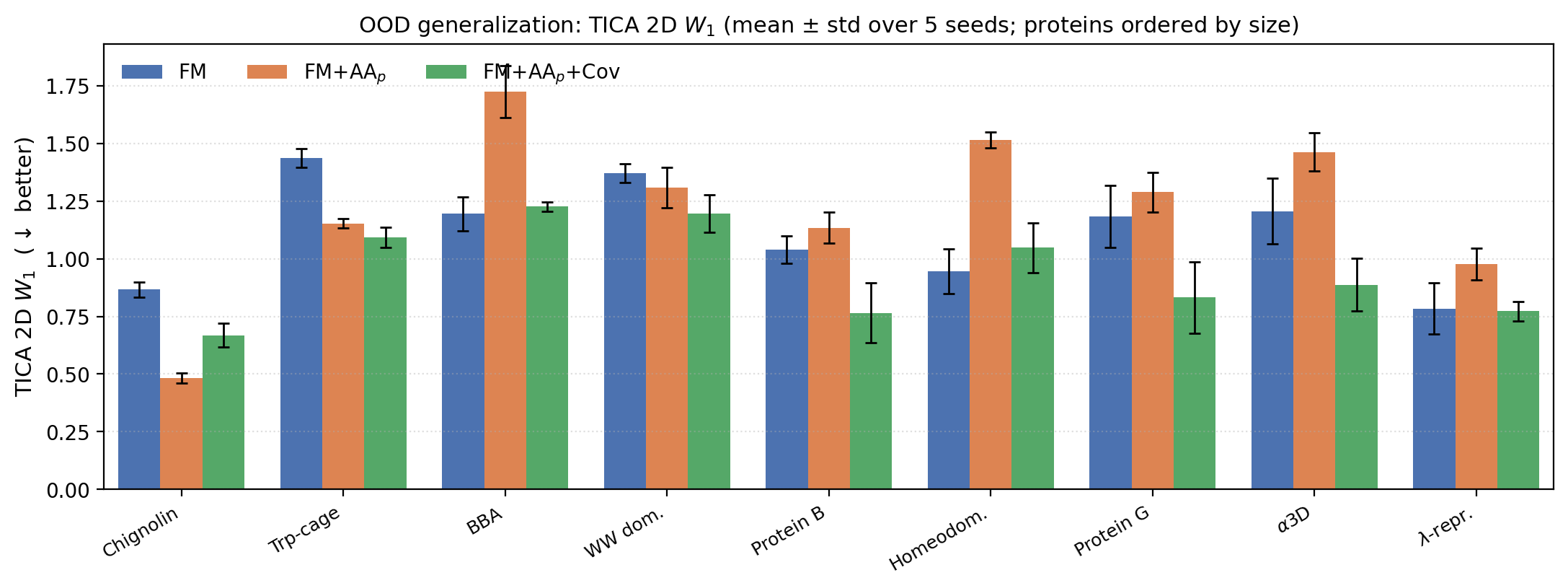}
    \caption{Out-of-distribution TICA~2D W1 distance. Grouped bars show the Wasserstein-1 distance over the joint distribution of the two slowest components (TIC0 and TIC1) for each variant (FM, FM+AA$_\text{p}$, FM+AA$_\text{p}$+Cov), mean over 5 seeds with whiskers denoting one standard deviation, for models trained on the single-chain dataset and evaluated on the nine benchmark proteins. Proteins are ordered by size (bead count) from left to right; lower is better.}
    \label{fig:ood_tica_2d_errors}
\end{figure}

Lastly, we show two of the slow-mode metrics---the TIC0 KL divergence (Figure \ref{fig:ood_tic0_kl_errors}) and TICA 2D W1 distance (Figure \ref{fig:ood_tica_2d_errors})---with their uncertainties with box-and-whiskers plots.

\end{document}